\documentclass{article}

\usepackage{arxiv}

\usepackage{graphicx}
\usepackage{subcaption}
\usepackage{booktabs}
\usepackage{microtype}
\usepackage[authoryear]{natbib}
\usepackage{placeins}

\usepackage{algorithm}
\usepackage{algpseudocode}

\usepackage{amsmath,amsfonts,bm}

\def\eqref#1{equation~\ref{#1}}

\def\1{\bm{1}}

\DeclareMathAlphabet{\mathsfit}{\encodingdefault}{\sfdefault}{m}{sl}
\SetMathAlphabet{\mathsfit}{bold}{\encodingdefault}{\sfdefault}{bx}{n}

\DeclareMathOperator*{\argmin}{arg\,min}

\usepackage{amsthm}
\newtheorem{theorem}{Theorem}

\newtheorem{lemma}{Lemma}

\usepackage[pdfusetitle]{hyperref}

\title{Deep Positive-Unlabeled Anomaly Detection for Contaminated Unlabeled Data}
\author{Hiroshi Takahashi, Tomoharu Iwata, Atsutoshi Kumagai, Yuuki Yamanaka}
\affiliation{NTT, Inc., Japan}
\correspondence{\href{mailto:hiroshibm.takahashi@ntt.com}{hiroshibm.takahashi@ntt.com}}
\codeurl{https://github.com/takahashihiroshi/pusvdd}
\manuscriptlicense{\href{https://creativecommons.org/licenses/by-nc-nd/4.0/}{CC BY-NC-ND 4.0}}
\renewcommand{\shorttitle}{Deep Positive-Unlabeled Anomaly Detection}
\begin{document}

\begin{abstract}
Semi-supervised anomaly detection, which aims to improve the anomaly detection performance by using a small amount of labeled anomaly data in addition to unlabeled data, has attracted attention.
Existing semi-supervised approaches assume that most unlabeled data are normal,
and train anomaly detectors by minimizing the anomaly scores for the unlabeled data
while maximizing those for the labeled anomaly data.
However, in practice, the unlabeled data are often contaminated with anomalies.
This weakens the effect of maximizing the anomaly scores for anomalies,
and prevents us from improving the detection performance.
To solve this,
we propose the deep positive-unlabeled anomaly detection framework,
which integrates positive-unlabeled learning with deep anomaly detection models such as autoencoders and deep support vector data descriptions.
Our approach enables the approximation of anomaly scores for normal data using the unlabeled data and the labeled anomaly data.
Therefore,
without labeled normal data,
our approach can train anomaly detectors by minimizing the anomaly scores for normal data while maximizing those for the labeled anomaly data.
We also provide a theoretical analysis establishing a generalization error bound for the proposed objective, guaranteeing that the empirical minimizer converges asymptotically to the ideal minimizer.
Our approach achieves better detection performance than existing approaches on various datasets.
\end{abstract}

\maketitle

\section{Introduction}
\label{sec:introduction}

Anomaly detection,
which aims to identify unusual data points,
is an important task in machine learning \citep{ruff2021unifying}.
It has been performed in various fields such as cyber-security \citep{kwon2019survey},
infrastructure monitoring \citep{borghesi2019anomaly},
novelty detection \citep{marchi2015novel},
medical diagnosis \citep{litjens2017survey},
and natural sciences \citep{min2017deep,cerri2019variational,pracht2020automated}.

In general, anomaly detection is performed by unsupervised learning because it does not require expensive and time-consuming labeling.
Unsupervised approaches assume that most unlabeled data are normal,
and try to detect anomalies by using an anomaly score, which represents the difference from normal data \citep{hinton2006reducing,ruff2018deep}.
Although these approaches are easy to handle,
their detection performance is limited because they cannot use information about anomalies.
To improve the detection performance,
semi-supervised anomaly detection uses a small amount of labeled anomaly data in addition to unlabeled data.
Existing semi-supervised approaches train anomaly detectors to minimize the anomaly scores for the unlabeled data,
and to maximize those for the labeled anomaly data \citep{hendrycks2018deep,ruff2019deep,yamanaka2019autoencoding}.
However, in practice, the unlabeled data are often contaminated with anomalies.
This weakens the effect of maximizing the anomaly scores for anomalies,
and prevents us from improving the anomaly detection performance.
This frequently occurs because it is difficult to label all anomalies.

To handle contaminated unlabeled data,
we propose the deep positive-unlabeled anomaly detection framework,
which integrates positive-unlabeled (PU) learning \citep{du2014analysis,du2015convex,kiryo2017positive} with deep anomaly detectors such as the autoencoder (AE) \citep{hinton2006reducing} and the deep support vector data description (DeepSVDD) \citep{ruff2018deep}.
PU learning assumes that an unlabeled data distribution is a mixture of normal and anomaly data distributions\footnote{
  Note that the anomaly data in the training dataset follow the anomaly data distribution,
  but new types of anomalies, unseen during training, may \emph{not} follow this distribution.
  In general, no distribution can fully represent all possible anomalies.
}.
Accordingly, the normal data distribution is approximated by using the unlabeled and anomaly data distributions.
With this assumption,
we approximate the anomaly scores for normal data using the unlabeled data and the labeled anomaly data.
Therefore,
without labeled normal data,
we can train anomaly detectors to minimize the anomaly scores for normal data,
and to maximize those for the labeled anomaly data.
Compared to existing semi-supervised approaches designed to handle contaminated unlabeled data \citep{zhang2018boosting,ju2020pumad,zhang2021elite,li2023deep,perini2023learning,shou2025read,durani2025weakly},
our approach is theoretically justified from the perspective of unbiased PU learning \citep{du2014analysis,du2015convex,kiryo2017positive}.

\begin{figure}
  \centering
    \begin{minipage}{0.245\linewidth}
      \includegraphics[width=\linewidth]{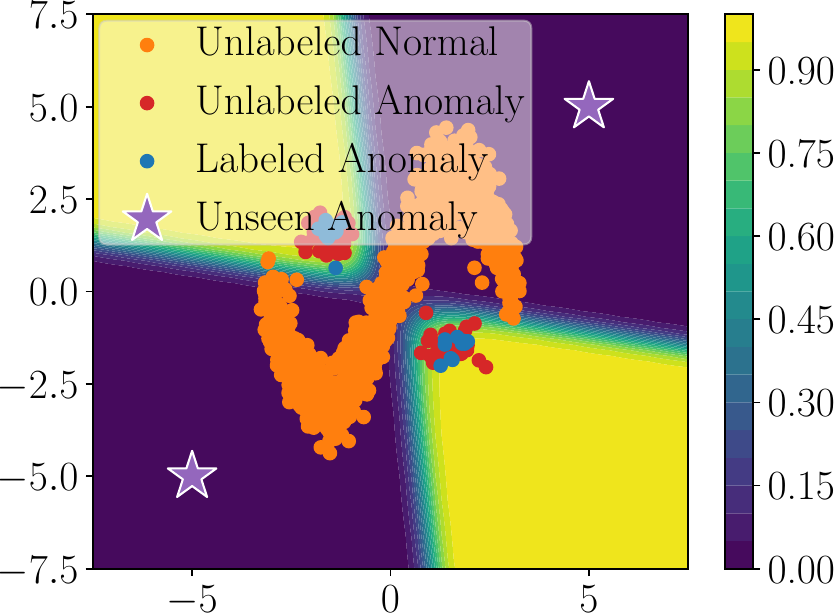}
      \subcaption{PU learning}
      \label{fig:toy_PU}
    \end{minipage}
    \begin{minipage}{0.245\linewidth}
      \includegraphics[width=\linewidth]{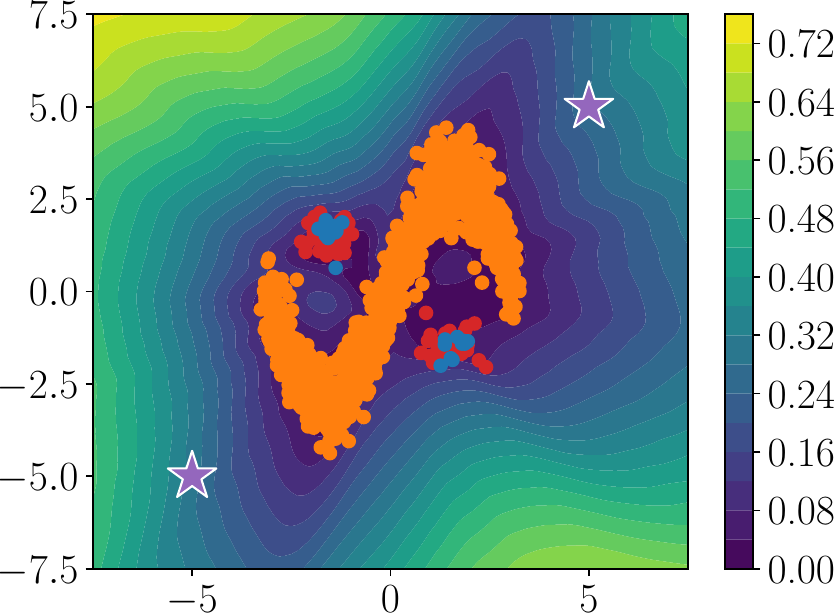}
      \subcaption{Unsupervised}
      \label{fig:toy_AE}
    \end{minipage}
    \begin{minipage}{0.245\linewidth}
      \includegraphics[width=\linewidth]{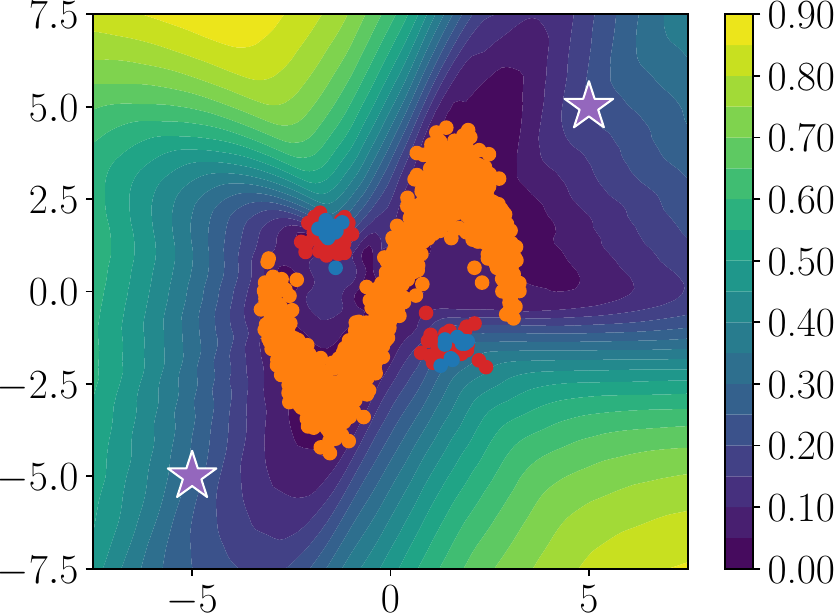}
      \subcaption{Semi-supervised}
      \label{fig:toy_ABC}
    \end{minipage}
    \begin{minipage}{0.245\linewidth}
      \includegraphics[width=\linewidth]{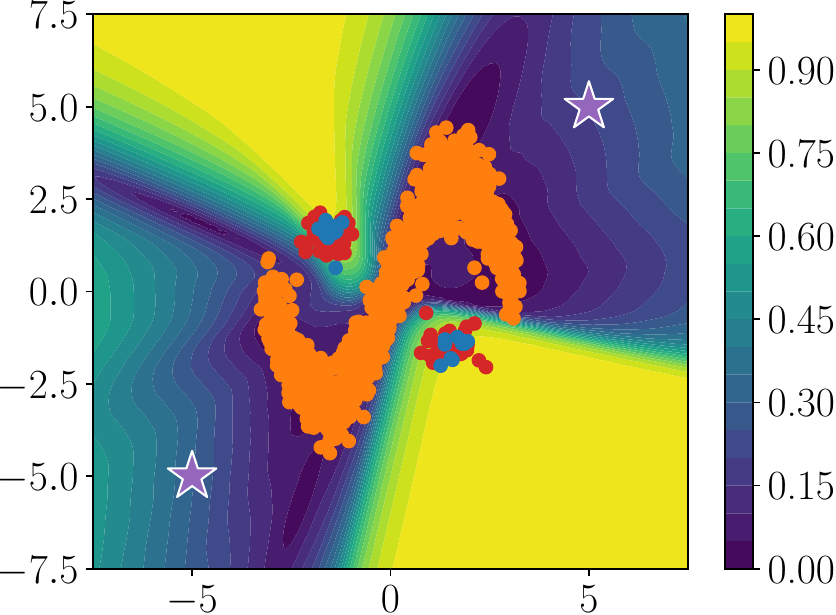}
      \subcaption{Proposed}
      \label{fig:toy_PUAE}
    \end{minipage}
  \caption{
    The comparison of the PU learning,
    the unsupervised anomaly detector (DAE),
    the semi-supervised anomaly detector (ABC),
    and the proposed method on the toy dataset.
    The unlabeled data in this dataset include both normal and anomaly data points.
    The yellow and blue in the contour maps represent abnormality and normality, respectively.
    The purple stars represent the examples of unseen anomalies, which are new types of anomalies unseen during training.
  }
  \label{fig:toy}
\end{figure}

Figure~\ref{fig:toy} compares the PU learning \citep{kiryo2017positive},
the unsupervised detector,
the semi-supervised detector,
and our approach on the toy dataset.
We used the denoising AE (DAE) \citep{vincent2008extracting} for the unsupervised detector,
and the autoencoding binary classifier (ABC) \citep{yamanaka2019autoencoding} for the semi-supervised detector.
Our approach is based on the DAE.
The toy dataset consists of unlabeled data and labeled anomaly data,
where the unlabeled data include both normal and anomaly data points.

We first focus on the PU learning,
which aims to train the binary classifier from the unlabeled data and the labeled anomaly data.
This can detect \textit{seen anomalies}, which are similar to anomalies included in the training dataset.
However, since its decision boundary is between normal data points and seen anomalies,
it cannot detect \textit{unseen anomalies}, which are new types of anomalies unseen during training,
such as novel anomalies and zero-day attacks \citep{wang2013k,pang2021toward,ding2022catching}.
We next focus on unsupervised and semi-supervised detectors.
They can detect unseen anomalies to some extent since they try to detect anomalies by using the difference from normal data.
The unsupervised detector cannot detect seen anomalies since it cannot use information about anomalies.
The semi-supervised detector can detect seen anomalies to some extent since it can use the labeled anomaly data.
However,
the contaminated dataset weakens the effect of maximizing the anomaly scores for anomalies in the semi-supervised detector.
Finally, we focus on our approach.
Our approach can detect seen anomalies according to the effectiveness of PU learning,
and can detect unseen anomalies to some extent according to the effectiveness of the deep anomaly detector.

Our framework is applicable to various anomaly detectors.
When selecting a detector,
we require that its loss function be non-negative and differentiable.
In this paper,
we apply our framework to the AE and the DeepSVDD.
We refer to the former as the positive-unlabeled autoencoder (PUAE),
and the latter as the positive-unlabeled support vector data description (PUSVDD).
We also provide a theoretical analysis of the proposed objective in Section~\ref{sec:theory},
establishing a generalization error bound that guarantees the empirical minimizer converges asymptotically to the ideal minimizer of the Positive-Negative (PN) risk, which can be computed if we have access to the normal data distribution, as the sample sizes grow.

Our contributions can be summarized as follows:
\begin{itemize}
\item To handle contaminated unlabeled data,
we propose the deep positive-unlabeled anomaly detection framework,
which integrates unbiased PU learning with deep anomaly detectors such as the AE and the DeepSVDD.
\item We theoretically justify the proposed objective via a Rademacher complexity-based generalization error bound (Section~\ref{sec:theory}).
\item We demonstrate that our approach outperforms existing approaches including state-of-the-art approaches \citep{li2023deep,durani2025weakly} on various datasets.
\end{itemize}

\section{Related Work}
\label{sec:related}

\subsection{Unsupervised Anomaly Detection}

Numerous unsupervised approaches have been presented,
ranging from shallow approaches such as the one-class support vector machine (OCSVM) \citep{tax2004support} and the isolation forest (IF) \citep{liu2008isolation}
to deep approaches such as the AE \citep{hinton2006reducing} and the DeepSVDD \citep{ruff2018deep}.
In addition, generative models such as the variational autoencoder \citep{kingma2014auto,kingma2015variational} and the generative adversarial nets \citep{goodfellow2014generative}
are also used for anomaly detection \citep{choi2018waic,serra2019input,ren2019likelihood,perera2019ocgan,xiao2020likelihood,havtorn2021hierarchical,yoon2021autoencoding}.
Although they are often used in anomaly detection,
their detection performance is limited because they cannot use information about anomalies.
For example, generative models may fail to detect anomalies that are obvious to the human eye \citep{nalisnick2018deep}.
Furthermore, these approaches assume that unlabeled data are mostly normal.
However, they are contaminated with anomalies in practice,
degrading the detection performance.
Several unsupervised approaches have been presented to handle such contaminated unlabeled data \citep{zhou2017anomaly,qiu2022latent,shang2023core}.
Among them, the latent outlier exposure (LOE) \citep{qiu2022latent} is a representative approach.
The LOE introduces the label for each data point as the latent variable,
and alternates between inferring the latent label and optimizing the parameter of the base anomaly detector.
Compared to these approaches,
our approach can achieve better detection performance by using the unlabeled data and the labeled anomaly data,
even if the unlabeled data are contaminated with anomalies.
We use the AE and the DeepSVDD as the base detectors for our approach.
In addition,
our approach can also be applied to the LOE by substituting its objective function into our proposed objective (Section~\ref{sec:puae}).

\subsection{Semi-supervised Anomaly Detection}

Several semi-supervised approaches have been presented,
aiming to improve the anomaly detection performance using labeled anomaly data in addition to unlabeled data \citep{hendrycks2018deep,yamanaka2019autoencoding,ruff2019deep,pang2023deep,zhou2021feature}.
Compared to these approaches,
our approach can effectively handle the unlabeled data that are contaminated with anomalies,
as described in Section~\ref{sec:puae}.

To handle contaminated unlabeled data,
several semi-supervised approaches have been presented,
including PU learning approaches \citep{zhang2018boosting,ju2020pumad,zhang2021elite,li2023deep,perini2023learning}.
Among them,
the semi-supervised outlier exposure with a limited labeling budget (SOEL) \citep{li2023deep} is a state-of-the-art approach.
The SOEL is a semi-supervised extension of the LOE \citep{qiu2022latent},
and presents the query strategy for the LOE, deciding which data should be labeled.
Our approach achieves equal to or better performance than SOEL on various datasets (Section~\ref{sec:experiments}).
We also conduct additional comparisons with the more recent WSAD-DT \citep{durani2025weakly} in Appendix~\ref{sec:additional_mvtec}, where our method outperforms it on MVTec AD.
The relationships with other recently proposed methods are as follows.
PReNet \citep{pang2023deep} is a weakly supervised approach targeting both seen and unseen anomalies; however, as explicitly stated in \citet{pang2023deep}, it focuses on multidimensional (tabular) data and is evaluated exclusively on tabular datasets. \citet{durani2025weakly} demonstrate that WSAD-DT outperforms PReNet, providing an indirect comparison with our method.
FEAWAD \citep{zhou2021feature} encodes features via autoencoders for weakly supervised detection, but is designed for tabular data and evaluated exclusively on tabular datasets; direct adaptation to image-based anomaly detection requires a non-trivial feature extraction protocol.
READ \citep{shou2025read} explicitly addresses contamination under limited supervision via robust statistics, yet is evaluated on tabular datasets and differs fundamentally from our PU learning-based approach.
WSAD-DT \citep{durani2025weakly} is a recent semi-supervised approach that, however, assumes unlabeled data are almost entirely normal.
Compared to SOEL, WSAD-DT, PReNet, FEAWAD, and READ,
our approach is theoretically justified from the perspective of unbiased PU learning \citep{du2014analysis,du2015convex,kiryo2017positive} and is designed for image-based anomaly detection with contaminated unlabeled data.

\subsection{Positive-Unlabeled Learning}

A lot of PU learning approaches have been presented for binary classification \citep{elkan2008learning,du2014analysis,du2015convex,kiryo2017positive,bekker2020learning,nakajima2023positive}.
Among them, our approach is based on the unbiased PU learning \citep{du2014analysis,du2015convex,kiryo2017positive}.
The empirical risk estimators of unbiased PU learning are consistent with respect to all popular loss functions, converging to the true PU risk as the dataset sizes grow (see Section~\ref{sec:puae}).
It is known that if the model is linear,
a particular loss function will result in convex optimization,
and the globally optimal solution can be obtained \citep{natarajan2013learning,patrini2016loss,niu2016theoretical}.
Despite this ideal property, when the model is complex such as neural networks,
the empirical risk estimator can become negative, potentially leading to the meaningless solution.
To address this issue,
following \citep{hammoudeh2020learning},
we take its absolute value, as described in Section~\ref{sec:puae}.
Non-negative risk estimators such as \citet{kiryo2017positive} and \citet{hammoudeh2020learning} remain the standard foundation for PU binary classification;
recent advances primarily address applications to specific settings such as class-prior shift \citep{nakajima2023positive} and distribution shift \citep{kumagai2025importance}, rather than replacing this core framework.
Compared to conventional PU learning that is based on the binary classifier,
our approach is based on deep anomaly detection models such as the AE and the DeepSVDD.
Although conventional PU learning cannot detect unseen anomalies since its decision boundary is between normal data points and seen anomalies,
our approach can detect both seen and unseen anomalies.

\section{Preliminaries}
\label{sec:preliminaries}

In this section, we first explain our problem setup.
Next, we review the AE \citep{hinton2006reducing} and the ABC \citep{yamanaka2019autoencoding}.
They are typical unsupervised and semi-supervised anomaly detection approaches,
and our framework can be applied to them.

\subsection{Problem Setup}
\label{sec:problem}

We can use unlabeled dataset $\mathcal{U}=\{\mathbf{x}_{1},\ldots,\mathbf{x}_{N}\}$ and anomaly dataset $\mathcal{A}=\{\tilde{\mathbf{x}}_{1},\ldots,\tilde{\mathbf{x}}_{M}\}$ for training.
$\mathcal{U}$ contains both normal data points and seen anomalies that are similar to those in $\mathcal{A}$.
The test dataset contains normal data points, seen anomalies, and unseen anomalies that are new types of anomalies unseen during training.
Our goal is to obtain a high-performance detector with $\mathcal{U}$ and $\mathcal{A}$.
Table~\ref{tab:terminology} summarizes these concepts and their correspondence to the training and test sets; Figure~\ref{fig:dataset_example} in Appendix~\ref{sec:dataset_example} illustrates a concrete example using MNIST.

\begin{table}[tb]
  \centering
  \caption{Key terminology and dataset correspondence.}
  \label{tab:terminology}
  \setlength{\tabcolsep}{3pt}
  \begin{tabular}{lll}
    \toprule
    Term & Definition & Split \\
    \midrule
    Unlabeled dataset ($\mathcal{U}$) & Contains normal data and seen anomalies without labels & Training \\
    Anomaly dataset ($\mathcal{A}$)   & Labeled seen anomaly samples                           & Training \\
    Normal data                       & Non-anomalous samples                                  & Training ($\mathcal{U}$) + Test \\
    Seen anomalies                    & Anomalies of the same types as those in $\mathcal{A}$ & Training ($\mathcal{A}$, $\mathcal{U}$) + Test \\
    Unseen anomalies                  & Anomalies of types not represented in $\mathcal{A}$   & Test only \\
    \bottomrule
  \end{tabular}
\end{table}

\subsection{Autoencoder}
\label{sec:ae}

For unsupervised anomaly detectors,
we train them only using the unlabeled dataset $\mathcal{U}$.
As an example, we focus on the AE,
which has been successfully applied to anomaly detection \citep{sakurada2014anomaly}.
The AE was originally presented for representation learning, which learns the representation of data points through data reconstruction.
Let $\mathbf{x}$ be a data point and $\mathbf{z}$ be its low-dimensional latent representation.
The AE consists of two neural networks:
encoder $E_{\theta}(\mathbf{x})$ and decoder $D_{\theta}(\mathbf{z})$,
where $\theta$ is the parameter of these neural networks.
$E_{\theta}(\mathbf{x})$ maps a data point $\mathbf{x}$ into a low-dimensional latent representation $\mathbf{z}$,
and $D_{\theta}(\mathbf{z})$ reconstructs the original data point $\mathbf{x}$ from the latent representation $\mathbf{z}$.
The reconstruction error for each data point $\mathbf{x}$ in the AE is defined as follows:
\begin{align}
\ell(\mathbf{x};\theta)=\left\Vert D_{\theta}(E_{\theta}(\mathbf{x}))-\mathbf{x}\right\Vert,
\label{eq:base_loss}
\end{align}
where $\left\Vert \cdot\right\Vert$ represents $\ell_{2}$ norm.
When the AE is used for anomaly detection,
all unlabeled data points are assumed to be normal.
We train the AE by minimizing the following objective function:
\begin{align}
\mathcal{L}_{\mathrm{AE}}(\theta)=\frac{1}{N}\sum_{n=1}^{N}\ell(\mathbf{x}_{n};\theta).
\end{align}
After training,
the AE is expected to successfully reconstruct normal data and fail to reconstruct anomaly data
because the training dataset is assumed to contain only normal data and no anomaly data.
Hence, the reconstruction error can be used for the anomaly score.

Although unsupervised approaches are widely used,
their detection performance is limited because they cannot use information about anomalies.

\subsection{Autoencoding Binary Classifier}
\label{sec:abc}

Semi-supervised anomaly detection aims to improve the anomaly detection performance using the unlabeled dataset $\mathcal{U}$ and the anomaly dataset $\mathcal{A}$.
A number of studies have been presented such as the ABC \citep{yamanaka2019autoencoding},
the deep semi-supervised anomaly detection (DeepSAD) \citep{ruff2019deep},
and the outlier exposure \citep{hendrycks2018deep}.
Here, we focus on the ABC, which is based on the AE.

Let $y=0$ be normal and $y=1$ be anomaly.
The ABC models the conditional probability of $y$ given $\mathbf{x}$ by using the reconstruction error $\ell(\mathbf{x};\theta)$ as follows:
\begin{align}
p_{\theta}(y|\mathbf{x})&=\begin{cases}
\exp(-\ell(\mathbf{x};\theta)) & (y=0)\\
1-\exp(-\ell(\mathbf{x};\theta)) & (y=1)
\end{cases}.
\end{align}
A small reconstruction error results in a higher probability of normality $p_{\theta}(y=0|\mathbf{x})$,
while a large reconstruction error results in a higher probability of abnormality $p_{\theta}(y=1|\mathbf{x})$.
With this conditional probability,
the ABC introduces the binary cross entropy as the loss function for each data point as follows:
\begin{align}
\ell_{\mathrm{BCE}}(\mathbf{x},y;\theta)
=-\log p_{\theta}(y|\mathbf{x})
=(1-y)\ell(\mathbf{x};\theta)-y\log(1-\exp(-\ell(\mathbf{x};\theta))).
\label{eq:bce_loss}
\end{align}
Like the AE, the ABC assumes all unlabeled data points to be normal.
The ABC is trained by minimizing the following objective function:
\begin{align}
\mathcal{L}_{\mathrm{ABC}}(\theta)=\frac{1}{N}\sum_{n=1}^{N}\ell_{\mathrm{BCE}}(\mathbf{x}_{n},0;\theta)
+\frac{1}{M}\sum_{m=1}^{M}\ell_{\mathrm{BCE}}(\tilde{\mathbf{x}}_{m},1;\theta).
\end{align}
This minimizes the reconstruction errors for the unlabeled data and maximizes those for the anomaly data.
Hence, after training,
the AE is expected to reconstruct the unlabeled data assumed to be normal,
and to fail to reconstruct anomalies.
Other semi-supervised anomaly detection approaches such as the DeepSAD \citep{ruff2019deep} and the outlier exposure \citep{hendrycks2018deep}
also minimize the anomaly scores for the unlabeled data and maximize those for the anomaly data.

However, the unlabeled dataset $\mathcal{U}$ is often contaminated with anomalies in practice.
This weakens the effect of maximizing the anomaly scores for anomalies,
and prevents us from improving the detection performance.
This frequently occurs because it is difficult to label all anomalies in the unlabeled dataset.

\section{Proposed Method}
\label{sec:proposed}

We aim to improve the detection performance even if the unlabeled dataset $\mathcal{U}$ contains anomalies.
To handle contaminated unlabeled data,
we propose the deep positive-unlabeled anomaly detection framework,
which integrates PU learning \citep{du2014analysis,du2015convex,kiryo2017positive} with deep anomaly detection models such as the AE and the DeepSVDD.
We refer to the former as the positive-unlabeled autoencoder (PUAE),
and the latter as the positive-unlabeled support vector data description (PUSVDD).
We also refer to anomalies as positive (+) samples, and normal data points as negative (-) samples.

\subsection{Positive-Unlabeled Autoencoder}
\label{sec:puae}

First, we explain the PUAE.
Let $p_{\mathcal{N}}$ be the normal data distribution,
$p_{\mathcal{A}}$ be the seen anomaly distribution,
and $p_{\mathcal{U}}$ be the unlabeled data distribution.
We assume that the datasets $\mathcal{U}$ and $\mathcal{A}$ are drawn from $p_{\mathcal{U}}$ and $p_{\mathcal{A}}$, respectively.
We also assume that $p_{\mathcal{U}}$ can be rewritten as follows:
\begin{align}
p_{\mathcal{U}}(\mathbf{x})=\alpha p_{\mathcal{A}}(\mathbf{x})+(1-\alpha)p_{\mathcal{N}}(\mathbf{x}),
\end{align}
where $\alpha\in[0,1]$ is the probability of anomaly occurrence in the unlabeled data.
Hence, $p_{\mathcal{N}}$ can be rewritten as follows:
\begin{align}
(1-\alpha)p_{\mathcal{N}}(\mathbf{x})=p_{\mathcal{U}}(\mathbf{x})-\alpha p_{\mathcal{A}}(\mathbf{x}).
\label{eq:p_normal}
\end{align}
Although $\alpha$ is a hyperparameter and is assumed to be known throughout this paper,
it can be estimated from the datasets $\mathcal{U}$ and $\mathcal{A}$ in conventional PU learning approaches \citep{menon2015learning,ramaswamy2016mixture,jain2016estimating,christoffel2016class}.

If we have access to the normal data distribution $p_{\mathcal{N}}$,
we can train the AE by minimizing the ideal objective function as follows:
\begin{align}
\mathcal{L}_{\mathrm{PN}}(\theta)=\alpha\mathbb{E}_{p_{\mathcal{A}}}[\ell_{\mathrm{BCE}}(\mathbf{x},1;\theta)]
+(1-\alpha)\mathbb{E}_{p_{\mathcal{N}}}[\ell_{\mathrm{BCE}}(\mathbf{x},0;\theta)],
\label{eq:pn_loss}
\end{align}
where $\mathbb{E}[\cdot]$ is the expectation.
Since we cannot access $p_{\mathcal{N}}$ in practice,
we have to approximate the second term in Eq.~(\ref{eq:pn_loss}).
According to Eq.~(\ref{eq:p_normal}),
this can be rewritten as follows:
\begin{align}
(1-\alpha)\mathbb{E}_{p_{\mathcal{N}}}[\ell_{\mathrm{BCE}}(\mathbf{x},0;\theta)]
=\mathbb{E}_{p_{\mathcal{U}}}[\ell_{\mathrm{BCE}}(\mathbf{x},0;\theta)]
-\alpha\mathbb{E}_{p_{\mathcal{A}}}[\ell_{\mathrm{BCE}}(\mathbf{x},0;\theta)].
\end{align}
Hence, by using the seen anomaly distribution $p_{\mathcal{A}}$ and the unlabeled data distribution $p_{\mathcal{U}}$,
$\mathcal{L}_{\mathrm{PN}}(\theta)$ can be rewritten as follows:
\begin{align}
\mathcal{L}_{\mathrm{PN}}(\theta)
=\alpha\mathbb{E}_{p_{\mathcal{A}}}[\ell_{\mathrm{BCE}}(\mathbf{x},1;\theta)]
+\mathbb{E}_{p_{\mathcal{U}}}[\ell_{\mathrm{BCE}}(\mathbf{x},0;\theta)]
-\alpha\mathbb{E}_{p_{\mathcal{A}}}[\ell_{\mathrm{BCE}}(\mathbf{x},0;\theta)].
\end{align}

With the datasets $\mathcal{U}$ and $\mathcal{A}$,
we can approximate $\mathcal{L}_{\mathrm{PN}}(\theta)$ by the empirical distribution as follows:
\begin{align}
\mathcal{L}_{\mathrm{PN}}(\theta)
\simeq\alpha\underbrace{\frac{1}{M}\sum_{m=1}^{M}\ell_{\mathrm{BCE}}(\tilde{\mathbf{x}}_{m},1;\theta)}_{\mathcal{L}_{\mathcal{A}}^{+}(\theta)}
+\underbrace{\frac{1}{N}\sum_{n=1}^{N}\ell_{\mathrm{BCE}}(\mathbf{x}_{n},0;\theta)}_{\mathcal{L}_{\mathcal{U}}^{-}(\theta)}
-\alpha\underbrace{\frac{1}{M}\sum_{m=1}^{M}\ell_{\mathrm{BCE}}(\tilde{\mathbf{x}}_{m},0;\theta)}_{\mathcal{L}_{\mathcal{A}}^{-}(\theta)}.
\label{eq:pn_loss_approx}
\end{align}
In this equation, the sum of the second and third terms is the approximation of the anomaly scores for normal data:
\begin{align}
(1-\alpha)\mathbb{E}_{p_{\mathcal{N}}}[\ell_{\mathrm{BCE}}(\mathbf{x},0;\theta)]\simeq\mathcal{L}_{\mathcal{U}}^{-}(\theta)-\alpha\mathcal{L}_{\mathcal{A}}^{-}(\theta).
\label{eq:normal_loss}
\end{align}
The left-hand side in Eq.~(\ref{eq:normal_loss}) is always greater than or equal to zero,
but the right-hand side can be negative.
In experiments, it often converges towards negative infinity, resulting in the meaningless solution.
To avoid this,
based on \citep{hammoudeh2020learning},
our training objective function to be minimized ensures that $\mathcal{L}_{\mathcal{U}}^{-}(\theta)-\alpha\mathcal{L}_{\mathcal{A}}^{-}(\theta)$ is not negative as follows:
\begin{align}
\mathcal{L}_{\mathrm{Proposed}}(\theta)=\alpha\mathcal{L}_{\mathcal{A}}^{+}(\theta)+\left|\mathcal{L}_{\mathcal{U}}^{-}(\theta)-\alpha\mathcal{L}_{\mathcal{A}}^{-}(\theta)\right|.
\label{eq:proposed}
\end{align}
The alternative approach of \citet{kiryo2017positive} uses $\max(0,\cdot)$ truncation;
in practice, when $\mathcal{L}_{\mathcal{U}}^{-}(\theta)-\alpha\mathcal{L}_{\mathcal{A}}^{-}(\theta)$ is negative,
\citet{kiryo2017positive} flips its sign, and minimizing it drives it back toward zero.
In our preliminary experiments, this caused the loss to spike at sign transitions, leading to unstable training.
The absolute value correction can avoid this,
and \citet{hammoudeh2020learning} prove that it yields a statistically consistent estimator of the true risk for any bounded loss function.
Our generalization error analysis in Section~\ref{sec:theory} specifically covers Eq.~(\ref{eq:proposed}),
establishing that the empirical minimizer of Eq.~(\ref{eq:proposed}) converges asymptotically to the ideal minimizer of the true PN risk.
We can optimize this training objective function by using the stochastic gradient descent (SGD) such as Adam \citep{kingma2015adam}.
We refer to this approach as the PUAE.
Algorithm~\ref{alg:proposed} in Appendix~\ref{sec:pseudo_code} shows the pseudo code of the PUAE.

\subsection{Positive-Unlabeled Support Vector Data Description}
\label{sec:pusvdd}

We next apply our framework to the DeepSVDD.
The DeepSVDD aims to pull the representation of the normal data towards the pre-defined center,
and push those of the anomaly data away from the center.
Let $f_{\theta}(\mathbf{x})$ be the feature extractor, like the encoder in the AE.
The loss function for each data point of the DeepSVDD is defined as follows:
\begin{align}
\tilde{\ell}(\mathbf{x};\theta)=\left\Vert f_{\theta}(\mathbf{x})-\mathbf{c}\right\Vert ^{2},
\label{eq:deep_svdd}
\end{align}
where $\mathbf{c}\neq\mathbf{0}$ is the pre-defined center vector, initialized as the mean of the encoder outputs over the training data after an initial pre-training phase of the AE.

The DeepSAD \citep{ruff2019deep} is a semi-supervised extension of the DeepSVDD.
The DeepSAD trains the DeepSVDD model to minimize Eq.~(\ref{eq:deep_svdd}) for the unlabeled data,
and to maximize it for the anomaly data.
The loss function for each data point of the DeepSAD is defined as follows:
\begin{align}
\tilde{\ell}_{\mathrm{SAD}}(\mathbf{x},y;\theta)=(1-y)\tilde{\ell}(\mathbf{x};\theta)+\frac{y}{\tilde{\ell}(\mathbf{x};\theta)}.
\label{eq:deep_sad}
\end{align}

We can apply our framework to the DeepSVDD by replacing Eq.~(\ref{eq:bce_loss}) in the PUAE with Eq.~(\ref{eq:deep_sad}),
while keeping all other components identical to the PUAE.
We refer to this approach as the PUSVDD.
In this way,
our framework is applicable to various anomaly detectors by substituting the loss function of the desired model into Eq.~(\ref{eq:base_loss}) in the PUAE or Eq.~(\ref{eq:deep_svdd}) in the PUSVDD.
When selecting a detector,
we require that its loss function be non-negative and differentiable.

\section{Theoretical Analysis}
\label{sec:theory}

The proposed objective $\mathcal{L}_{\mathrm{Proposed}}(\theta)$ (Eq.~(\ref{eq:proposed})) is an empirical approximation of the ideal PN risk $\mathcal{L}_{\mathrm{PN}}(\theta)$ (Eq.~(\ref{eq:pn_loss})).
Minimizing $\mathcal{L}_{\mathrm{PN}}(\theta)$ would yield the ideal minimizer $\theta^{*}=\argmin_{\theta\in\Theta}\mathcal{L}_{\mathrm{PN}}(\theta)$, where $\Theta$ denotes the parameter space; in practice, however, we minimize $\mathcal{L}_{\mathrm{Proposed}}(\theta)$ over finite data, obtaining $\hat{\theta}=\argmin_{\theta\in\Theta}\mathcal{L}_{\mathrm{Proposed}}(\theta)$, which may deviate from $\theta^{*}$.
This section analyzes the generalization error of $\hat{\theta}$ and establishes that it converges asymptotically to $\theta^{*}$ as the sample sizes grow.
Theorem~\ref{thm:generalization_error} below adapts classical PU learning generalization bounds \citep{kiryo2017positive} to the representation-based anomaly detection losses used in the PUAE and PUSVDD.

Let $\mathcal{H}=\{f_\theta:\theta\in\Theta\}$ be the hypothesis class corresponding to $\Theta$.
For this $\mathcal{H}$,
we define the empirical Rademacher complexity $\hat{\mathcal{R}}_{K}^{q}(\mathcal{H})$ and the Rademacher complexity $\mathcal{R}_{K}^{q}(\mathcal{H})$ as follows:
\begin{align}
\hat{\mathcal{R}}_{K}^{q}(\mathcal{H})
=\mathbb{E}_{\boldsymbol{\sigma}}\!\left[\sup_{\theta\in\Theta}\left|\frac{1}{K}\sum_{k=1}^{K}\sigma_{k}f_{\theta}(\mathbf{x}_{k})\right|\right],
\quad
\mathcal{R}_{K}^{q}(\mathcal{H})
=\mathbb{E}_{\mathcal{D}}\!\left[\hat{\mathcal{R}}_{K}^{q}(\mathcal{H})\right],
\end{align}
where $\mathcal{D}=\{\mathbf{x}_k\}_{k=1}^K$ is a sample set drawn i.i.d.\ from $q$, and $\boldsymbol{\sigma}=(\sigma_k)_{k=1}^K\in\{-1,+1\}^K$ are independent Rademacher random variables.
The Rademacher complexity measures the expressiveness of $\mathcal{H}$ and is a standard tool in generalization analysis \citep{mohri2018foundations}.
We assume that $\ell_{\mathrm{BCE}}$ is bounded in $[0,B]$, that $\mathcal{U}$ and $\mathcal{A}$ are drawn i.i.d.\ and independently from $p_{\mathcal{U}}$ and $p_{\mathcal{A}}$, and that $p_{\mathcal{U}}=\alpha p_{\mathcal{A}}+(1-\alpha)p_{\mathcal{N}}$.
Let $C_1$ and $C_0$ denote the Lipschitz constants of $\ell_{\mathrm{BCE}}(\mathbf{x},1;\theta)$ and $\ell_{\mathrm{BCE}}(\mathbf{x},0;\theta)$ with respect to $f_\theta(\mathbf{x})$, respectively.

\begin{theorem}
\label{thm:generalization_error}
Let $\hat{\theta}=\argmin_{\theta\in\Theta}\mathcal{L}_{\mathrm{Proposed}}(\theta)$ be the minimizer obtained from the training dataset $\mathcal{D}_{tr}=\{\mathcal{U},\mathcal{A}\}$, and suppose $\mathcal{H}$ contains the true model $f_{\theta^{*}}$.
Then
\begin{align}
\mathbb{E}_{\mathcal{D}_{tr}}\!\left[\mathcal{L}_{\mathrm{PN}}(\hat{\theta})\right]
\leq\inf_{\theta\in\Theta}\mathcal{L}_{\mathrm{PN}}(\theta)
+8\left\{\alpha C_{1}\mathcal{R}_{M}^{p_{\mathcal{A}}}(\mathcal{H})+C_{0}\mathcal{R}_{N}^{p_{\mathcal{U}}}(\mathcal{H})+\alpha C_{0}\mathcal{R}_{M}^{p_{\mathcal{A}}}(\mathcal{H})\right\}.
\label{eq:generalization_error}
\end{align}
\end{theorem}

The proof of Theorem~\ref{thm:generalization_error} uses the following two lemmas.

\begin{lemma}
\label{lem:excess_risk}
Let $\hat{\theta}=\argmin_{\theta\in\Theta}\mathcal{L}_{\mathrm{Proposed}}(\theta)$.
Then
\begin{align}
\mathcal{L}_{\mathrm{PN}}(\hat{\theta})
\leq\inf_{\theta\in\Theta}\mathcal{L}_{\mathrm{PN}}(\theta)
+2\sup_{\theta\in\Theta}\bigl|\mathcal{L}_{\mathrm{PN}}(\theta)-\mathcal{L}_{\mathrm{Proposed}}(\theta)\bigr|.
\label{eq:excess_decomp}
\end{align}
\end{lemma}

\begin{proof}
Since $\hat{\theta}$ minimizes $\mathcal{L}_{\mathrm{Proposed}}$, for any $\theta\in\Theta$:
\begin{align}
\mathcal{L}_{\mathrm{PN}}(\hat{\theta})
&=\mathcal{L}_{\mathrm{Proposed}}(\hat{\theta})+\bigl(\mathcal{L}_{\mathrm{PN}}(\hat{\theta})-\mathcal{L}_{\mathrm{Proposed}}(\hat{\theta})\bigr)\nonumber\\
&\leq\mathcal{L}_{\mathrm{Proposed}}(\hat{\theta})+\sup_{\theta'\in\Theta}\bigl|\mathcal{L}_{\mathrm{PN}}(\theta')-\mathcal{L}_{\mathrm{Proposed}}(\theta')\bigr|,
\end{align}
and
\begin{align}
\mathcal{L}_{\mathrm{Proposed}}(\hat{\theta})
&\leq\mathcal{L}_{\mathrm{Proposed}}(\theta)\nonumber\\
&=\mathcal{L}_{\mathrm{PN}}(\theta)+\bigl(\mathcal{L}_{\mathrm{Proposed}}(\theta)-\mathcal{L}_{\mathrm{PN}}(\theta)\bigr)\nonumber\\
&\leq\mathcal{L}_{\mathrm{PN}}(\theta)+\sup_{\theta'\in\Theta}\bigl|\mathcal{L}_{\mathrm{Proposed}}(\theta')-\mathcal{L}_{\mathrm{PN}}(\theta')\bigr|.
\end{align}
Combining these two inequalities and taking the infimum over $\theta$ gives Eq.~(\ref{eq:excess_decomp}).
\end{proof}

\begin{lemma}
\label{lem:uniform_dev}
\begin{align}
\mathbb{E}_{\mathcal{D}_{tr}}\!\left[\sup_{\theta\in\Theta}\bigl|\mathcal{L}_{\mathrm{PN}}(\theta)-\mathcal{L}_{\mathrm{Proposed}}(\theta)\bigr|\right]
\leq 4\left\{\alpha C_{1}\mathcal{R}_{M}^{p_{\mathcal{A}}}(\mathcal{H})+C_{0}\mathcal{R}_{N}^{p_{\mathcal{U}}}(\mathcal{H})+\alpha C_{0}\mathcal{R}_{M}^{p_{\mathcal{A}}}(\mathcal{H})\right\}.
\label{eq:unif_dev}
\end{align}
\end{lemma}

\begin{proof}
We first rewrite $\mathcal{L}_{\mathrm{PN}}(\theta)$ using $(1-\alpha)p_{\mathcal{N}}=p_{\mathcal{U}}-\alpha p_{\mathcal{A}}$:
\begin{align}
\mathcal{L}_{\mathrm{PN}}(\theta)
&=\alpha\mathbb{E}_{p_{\mathcal{A}}}[\ell_{\mathrm{BCE}}(\mathbf{x},1;\theta)]+(1-\alpha)\mathbb{E}_{p_{\mathcal{N}}}[\ell_{\mathrm{BCE}}(\mathbf{x},0;\theta)]\nonumber\\
&=\alpha\mathbb{E}_{p_{\mathcal{A}}}[\ell_{\mathrm{BCE}}(\mathbf{x},1;\theta)]+\mathbb{E}_{p_{\mathcal{U}}}[\ell_{\mathrm{BCE}}(\mathbf{x},0;\theta)]-\alpha\mathbb{E}_{p_{\mathcal{A}}}[\ell_{\mathrm{BCE}}(\mathbf{x},0;\theta)]\nonumber\\
&=\alpha\mathbb{E}_{p_{\mathcal{A}}}[\ell_{\mathrm{BCE}}(\mathbf{x},1;\theta)]+\left|\mathbb{E}_{p_{\mathcal{U}}}[\ell_{\mathrm{BCE}}(\mathbf{x},0;\theta)]-\alpha\mathbb{E}_{p_{\mathcal{A}}}[\ell_{\mathrm{BCE}}(\mathbf{x},0;\theta)]\right|,
\end{align}
where the last equality holds because $\mathbb{E}_{p_{\mathcal{U}}}[\ell_{\mathrm{BCE}}(\mathbf{x},0;\theta)]-\alpha\mathbb{E}_{p_{\mathcal{A}}}[\ell_{\mathrm{BCE}}(\mathbf{x},0;\theta)]=(1-\alpha)\mathbb{E}_{p_{\mathcal{N}}}[\ell_{\mathrm{BCE}}(\mathbf{x},0;\theta)]\geq 0$, since $\ell_{\mathrm{BCE}}\in[0,B]$ and $\alpha\in[0,1]$.
The proposed objective takes the corresponding empirical form:
\begin{align}
\mathcal{L}_{\mathrm{Proposed}}(\theta)
=\frac{\alpha}{M}\sum_{m=1}^{M}\ell_{\mathrm{BCE}}(\tilde{\mathbf{x}}_{m},1;\theta)
+\left|\frac{1}{N}\sum_{n=1}^{N}\ell_{\mathrm{BCE}}(\mathbf{x}_{n},0;\theta)-\frac{\alpha}{M}\sum_{m=1}^{M}\ell_{\mathrm{BCE}}(\tilde{\mathbf{x}}_{m},0;\theta)\right|.
\end{align}
Applying the triangle inequality and $\bigl||a|-|b|\bigr|\leq|a-b|$ yields
\begin{align}
\sup_{\theta\in\Theta}\bigl|\mathcal{L}_{\mathrm{PN}}(\theta)-\mathcal{L}_{\mathrm{Proposed}}(\theta)\bigr|
\leq\alpha\Delta_{\mathcal{A}}^{(1)}+\Delta_{\mathcal{U}}^{(0)}+\alpha\Delta_{\mathcal{A}}^{(0)},
\end{align}
where
\begin{align}
\Delta_{\mathcal{A}}^{(1)}&=\sup_{\theta\in\Theta}\left|\mathbb{E}_{p_{\mathcal{A}}}[\ell_{\mathrm{BCE}}(\mathbf{x},1;\theta)]-\frac{1}{M}\sum_{m=1}^{M}\ell_{\mathrm{BCE}}(\tilde{\mathbf{x}}_{m},1;\theta)\right|,\\
\Delta_{\mathcal{U}}^{(0)}&=\sup_{\theta\in\Theta}\left|\mathbb{E}_{p_{\mathcal{U}}}[\ell_{\mathrm{BCE}}(\mathbf{x},0;\theta)]-\frac{1}{N}\sum_{n=1}^{N}\ell_{\mathrm{BCE}}(\mathbf{x}_{n},0;\theta)\right|,\\
\Delta_{\mathcal{A}}^{(0)}&=\sup_{\theta\in\Theta}\left|\mathbb{E}_{p_{\mathcal{A}}}[\ell_{\mathrm{BCE}}(\mathbf{x},0;\theta)]-\frac{1}{M}\sum_{m=1}^{M}\ell_{\mathrm{BCE}}(\tilde{\mathbf{x}}_{m},0;\theta)\right|.
\end{align}
By the symmetrization argument and Talagrand's contraction lemma \citep{mohri2018foundations}, we obtain the following bounds:
\begin{align}
\mathbb{E}_{\mathcal{D}_{tr}}\!\left[\Delta_{\mathcal{A}}^{(1)}\right]\leq 4C_{1}\mathcal{R}_{M}^{p_{\mathcal{A}}}(\mathcal{H}),\quad
\mathbb{E}_{\mathcal{D}_{tr}}\!\left[\Delta_{\mathcal{U}}^{(0)}\right]\leq 4C_{0}\mathcal{R}_{N}^{p_{\mathcal{U}}}(\mathcal{H}),\quad
\mathbb{E}_{\mathcal{D}_{tr}}\!\left[\Delta_{\mathcal{A}}^{(0)}\right]\leq 4C_{0}\mathcal{R}_{M}^{p_{\mathcal{A}}}(\mathcal{H}).
\end{align}
Taking the expectation and combining gives Eq.~(\ref{eq:unif_dev}).
\end{proof}

\begin{proof}[Proof of Theorem~\ref{thm:generalization_error}]
Taking the expectation of Lemma~\ref{lem:excess_risk} and applying Lemma~\ref{lem:uniform_dev}:
\begin{align}
\mathbb{E}_{\mathcal{D}_{tr}}\!\left[\mathcal{L}_{\mathrm{PN}}(\hat{\theta})\right]
&\leq\inf_{\theta\in\Theta}\mathcal{L}_{\mathrm{PN}}(\theta)+2\mathbb{E}_{\mathcal{D}_{tr}}\!\left[\sup_{\theta\in\Theta}\bigl|\mathcal{L}_{\mathrm{PN}}(\theta)-\mathcal{L}_{\mathrm{Proposed}}(\theta)\bigr|\right]\nonumber\\
&\leq\inf_{\theta\in\Theta}\mathcal{L}_{\mathrm{PN}}(\theta)+8\left\{\alpha C_{1}\mathcal{R}_{M}^{p_{\mathcal{A}}}(\mathcal{H})+C_{0}\mathcal{R}_{N}^{p_{\mathcal{U}}}(\mathcal{H})+\alpha C_{0}\mathcal{R}_{M}^{p_{\mathcal{A}}}(\mathcal{H})\right\}.\qedhere
\end{align}
\end{proof}

Since $\mathcal{R}_{K}^{q}(\mathcal{H})=\mathcal{O}(1/\sqrt{K})$ holds in general \citep{mohri2018foundations},
the second term on the right-hand side of Eq.~(\ref{eq:generalization_error}) is $\mathcal{O}(\alpha C_1/\sqrt{M}+C_0/\sqrt{N}+\alpha C_0/\sqrt{M})$.
Hence, as $N,M\to\infty$, this term vanishes and $\mathbb{E}[\mathcal{L}_{\mathrm{PN}}(\hat{\theta})]$ asymptotically approaches $\inf_{\theta\in\Theta}\mathcal{L}_{\mathrm{PN}}(\theta)$.
This confirms the asymptotic convergence of $\hat{\theta}$ to $\theta^{*}$ stated at the outset of this section.

\section{Experiments}
\label{sec:experiments}

\subsection{Data}

To evaluate our approach,
we used the following eight image datasets:
MNIST~\citep{salakhutdinov2008quantitative},
FashionMNIST~\citep{xiao2017fashion},
SVHN~\citep{netzer2011reading},
CIFAR10~\citep{krizhevsky2009learning},
CIFAR100~\citep{krizhevsky2009learning},
Path, OCT, and Tissue~\citep{yang2021medmnist,yang2023medmnist}.

First, we explain the first four datasets.
MNIST is the handwritten digits,
FashionMNIST is the fashion product images,
SVHN is the house number digits,
and CIFAR10 is the animal and vehicle images.
We resized all datasets to $32 \times 32$ resolution.
These datasets consist of 10 class images.
For MNIST and SVHN, we used the digits as class indices.
For FashionMNIST, we indexed labels as:
\{T-shirt/top: 0, Trouser: 1, Pullover: 2, Dress: 3, Coat: 4, Sandal: 5, Shirt: 6, Sneaker: 7, Bag: 8, and Ankle boot: 9\}.
For CIFAR10, we indexed labels as:
\{airplane: 0, automobile: 1, bird: 2, cat: 3, deer: 4, dog: 5, frog: 6, horse: 7, ship: 8, and truck: 9\}.
We extend the experiments in \citep{ruff2018deep} to semi-supervised anomaly detection with contaminated unlabeled data.
Of the 10 classes,
we used one class as normal,
another class as unseen anomaly,
and the remaining classes as seen anomaly.
For example, with MNIST,
if we use the digit 1 as normal and the digit 0 as unseen anomaly,
seen anomaly corresponds to the digits 2, 3, 4, 5, 6, 7, 8, and 9.
For all datasets, we used class 0 as unseen anomaly,
and selected one normal class from the remaining 9 classes.
The training dataset consists of 5,000 samples,
of which 4,500 samples are unlabeled normal data points,
250 samples are labeled seen anomalies,
and 250 samples are unlabeled seen anomalies.
That is,
the unlabeled data points in this dataset are contaminated with seen anomalies.
We used 10\% of the training dataset as the validation dataset for early stopping to prevent over-fitting \citep{goodfellow2016deep}.
The test dataset consists of 2,000 samples,
about half of which are normal and the rest are anomalies, including both seen and unseen anomalies.
More specifically,
normal data points are sampled from the normal class in the test dataset,
with a maximum of 1,000 samples,
while both seen and unseen anomalies are sampled from their respective classes,
with a maximum of 500 samples each.
The example of the MNIST dataset is provided in Appendix~\ref{sec:dataset_example}.

Next, we explain the last four datasets.
CIFAR100 is just like CIFAR10 but consists of 100 classes.
These classes are grouped into 20 superclasses,
from which we used nature-related classes as normal and human-related classes as anomalies.
We used the people class as unseen anomaly.
Path, OCT, and Tissue are real medical image datasets,
which are also used in \citep{li2023deep}.
Path is a colorectal cancer histology dataset with 9 tissue types.
OCT is a retinal optical coherence tomography dataset with 4 diagnostic categories.
Tissue is a kidney cortex cell dataset with 8 categories.
For Path and Tissue,
we used the first class as unseen anomaly,
selected one class from the remaining ones as normal,
and treated the rest as seen anomaly.
For OCT,
since a pre-defined normal class exists,
we used it as normal, used the first class as unseen anomaly,
and treated the remaining classes as seen anomaly.

\subsection{Methods}

For comparison with our PUAE and PUSVDD,
we used the following unsupervised and semi-supervised approaches.

\textbf{Unsupervised approaches:}
We used the IF \citep{liu2008isolation} as the shallow approach,
and used the AE \citep{hinton2006reducing} and the DeepSVDD \citep{ruff2018deep} as the deep approaches.
We also used the LOE \citep{qiu2022latent},
which is robust to the anomalies in the unlabeled data.
We chose the DeepSVDD as the base detector for the LOE.

\textbf{Semi-supervised approaches:}
We used the ABC \citep{yamanaka2019autoencoding}, the DeepSAD \citep{ruff2019deep},
and the SOEL \citep{li2023deep} that is a semi-supervised extension of the LOE.
We chose the DeepSVDD as the base detector for the SOEL.
We also used the PU learning binary classifier (PU) \citep{kiryo2017positive} for reference.

\subsection{Setup}
\label{sec:setup}

First, we outline the setups for all approaches except the IF.
We used convolutional neural networks for the AE, the DeepSVDD, and the PU.
The network architecture follows \citep{ruff2018deep}.
For the AE-based and DeepSVDD-based approaches,
we set the dimension of the latent variable to 128.
For the DeepSVDD-based approaches,
we used no bias terms in each layer,
pre-trained these feature extractors as the AE,
and set the center $\mathbf{c}$ in Eq.~(\ref{eq:deep_svdd}) to the mean of the outputs of the encoder.
We trained all methods by using Adam \citep{kingma2015adam} with a mini-batch size of 128.
We set the learning rate to $10^{-4}$ and the maximum number of epochs to 200.
We also used the weight decay \citep{goodfellow2016deep} with $10^{-3}$ and used early-stopping~\citep{goodfellow2016deep} based on the validation dataset.
We set $\alpha=0.1$, the probability of anomaly occurrence in the unlabeled data, for the PUAE, the PUSVDD, the PU, the LOE, and the SOEL.
Next, we outline the setup for the IF.
We used the scikit-learn implementation \citep{pedregosa2011scikit} and kept all hyperparameters at their default values in our experiments.

We trained unsupervised approaches using the unlabeled data\footnote{
  Note that we did not use the labeled anomaly data, since unsupervised approaches cannot effectively use them.
},
while we trained semi-supervised approaches using both the unlabeled data and the labeled anomaly data.
To measure the detection performance,
we calculated the AUROC scores for all datasets.
We ran all experiments five times while changing the random seeds.
The machine specifications used in the experiments are as follows:
the CPU is AMD EPYC 9124 16-Core Processor,
the memory size is 512GB,
and the GPU is NVIDIA RTX 6000 Ada.

\subsection{Results}

\begin{table}
\caption{
  Comparison of anomaly detection performance on MNIST, FashionMNIST, SVHN, and CIFAR10.
}
\centering
\begin{tabular}{lrrrr}
\toprule
& MNIST & FashionMNIST & SVHN & CIFAR10 \\
\midrule
IF & 0.885 $\pm$ 0.062 & 0.916 $\pm$ 0.077 & 0.501 $\pm$ 0.014 & 0.610 $\pm$ 0.095 \\
AE & 0.912 $\pm$ 0.042 & 0.841 $\pm$ 0.102 & 0.562 $\pm$ 0.040 & 0.535 $\pm$ 0.120 \\
DeepSVDD & 0.937 $\pm$ 0.045 & 0.921 $\pm$ 0.088 & 0.582 $\pm$ 0.035 & 0.709 $\pm$ 0.054 \\
LOE & 0.945 $\pm$ 0.033 & 0.916 $\pm$ 0.094 & 0.624 $\pm$ 0.054 & 0.718 $\pm$ 0.062 \\
\midrule
ABC & 0.916 $\pm$ 0.042 & 0.841 $\pm$ 0.104 & 0.562 $\pm$ 0.041 & 0.535 $\pm$ 0.119 \\
DeepSAD & 0.942 $\pm$ 0.041 & 0.928 $\pm$ 0.089 & 0.652 $\pm$ 0.034 & 0.726 $\pm$ 0.051 \\
SOEL & 0.965 $\pm$ 0.026 & 0.936 $\pm$ 0.079 & 0.727 $\pm$ 0.043 & 0.775 $\pm$ 0.057 \\
PU & 0.962 $\pm$ 0.034 & 0.922 $\pm$ 0.092 & 0.681 $\pm$ 0.096 & 0.693 $\pm$ 0.120 \\
\midrule
PUAE & 0.983 $\pm$ 0.015 & 0.918 $\pm$ 0.085 & 0.689 $\pm$ 0.060 & 0.667 $\pm$ 0.066 \\
PUSVDD & \textbf{0.989 $\pm$ 0.012} & \textbf{0.948 $\pm$ 0.079} & \textbf{0.747 $\pm$ 0.080} & \textbf{0.803 $\pm$ 0.046} \\
\bottomrule
\end{tabular}
\label{tab:results_first}
\end{table}

\begin{table}
\caption{
  Comparison of anomaly detection performance on CIFAR100, Path, OCT and Tissue.
}
\centering
\begin{tabular}{lrrrr}
\toprule
& CIFAR100 & Path & OCT & Tissue \\
\midrule
IF & 0.604 $\pm$ 0.004 & \textbf{0.809 $\pm$ 0.118} & 0.714 $\pm$ 0.004 & 0.472 $\pm$ 0.187 \\
AE & 0.589 $\pm$ 0.010 & 0.605 $\pm$ 0.240 & \textbf{0.860 $\pm$ 0.005} & 0.468 $\pm$ 0.178 \\
DeepSVDD & 0.587 $\pm$ 0.026 & 0.759 $\pm$ 0.148 & 0.726 $\pm$ 0.052 & 0.661 $\pm$ 0.055 \\
LOE & 0.576 $\pm$ 0.035 & 0.721 $\pm$ 0.160 & 0.783 $\pm$ 0.030 & 0.635 $\pm$ 0.086 \\
\midrule
ABC & 0.590 $\pm$ 0.010 & 0.604 $\pm$ 0.241 & \textbf{0.857 $\pm$ 0.001} & 0.472 $\pm$ 0.177 \\
DeepSAD & 0.594 $\pm$ 0.012 & 0.763 $\pm$ 0.187 & \textbf{0.823 $\pm$ 0.038} & 0.683 $\pm$ 0.053 \\
SOEL & \textbf{0.633 $\pm$ 0.013} & 0.791 $\pm$ 0.145 & \textbf{0.856 $\pm$ 0.016} & 0.703 $\pm$ 0.062 \\
PU & 0.541 $\pm$ 0.025 & \textbf{0.807 $\pm$ 0.132} & 0.614 $\pm$ 0.109 & 0.633 $\pm$ 0.082 \\
\midrule
PUAE & 0.623 $\pm$ 0.014 & \textbf{0.776 $\pm$ 0.168} & \textbf{0.847 $\pm$ 0.011} & 0.594 $\pm$ 0.104 \\
PUSVDD & \textbf{0.637 $\pm$ 0.017} & \textbf{0.831 $\pm$ 0.152} & \textbf{0.857 $\pm$ 0.017} & \textbf{0.731 $\pm$ 0.077} \\
\bottomrule
\end{tabular}
\label{tab:results_last}
\end{table}

Tables~\ref{tab:results_first} and \ref{tab:results_last} compare the detection performance on each dataset.
We show the average of the AUROC scores for all normal classes.
We used bold to highlight the best results and statistically non-different results according to a pair-wise $t$-test.
We used 5\% as the p-value.

First, we focus on unsupervised approaches.
The IF and the AE show significant performance variations across different datasets.
Although the IF performed well on Path and the AE performed well on OCT,
their performance became poor on other datasets.
The DeepSVDD outperformed the IF and the AE,
and the LOE, which is based on the DeepSVDD, often performed better than the DeepSVDD.
However, since the LOE estimates anomalies within the contaminated unlabeled data in an unsupervised manner,
incorrect estimations can lead to degraded performance.

Next, we focus on semi-supervised approaches.
The ABC and the DeepSAD performed equal to or slightly better than the AE and the DeepSVDD, respectively.
The reason for this is that ABC and DeepSAD assume that the unlabeled data are not contaminated with anomalies,
which weakens the effect of maximizing anomaly scores for the labeled anomaly data.
On the other hand, the SOEL outperformed DeepSAD in all cases.
This is because the SOEL is capable of handling anomalies present in the unlabeled data.

Finally, we focus on the proposed methods.
In most cases,
the PUAE and the PUSVDD performed better than the AE and the ABC, the DeepSVDD and the DeepSAD, respectively.
Especially, the PUSVDD achieved the best performance across all datasets.
These results strongly indicate the effectiveness of our framework,
which integrates PU learning with deep anomaly detection models.
The detection performance under varying numbers of unlabeled anomalies are shown in Appendix~\ref{sec:additional_n_unlabeled_anomalies},
where our PUSVDD achieves performance equal to or better than that of SOEL.
Per-class AUROC results for PUSVDD and SOEL on MNIST, FashionMNIST, SVHN, CIFAR10, Path, and Tissue are provided in Appendix~\ref{sec:per_class}.

We also focus on the difference between the proposed methods and PU.
The proposed methods performed equal to or better than the PU in all datasets.
The reason is as follows.
Since the PU is for binary classification,
it sets the decision boundary between normal data points and seen anomalies.
This prevents us from detecting unseen anomalies.
On the other hand,
our approach can detect unseen anomalies since it can model normal data points by the anomaly detector.
The detection performance for seen and unseen anomalies are shown in Appendix~\ref{sec:additional_each}.
We also show the additional experiments using the MVTec Anomaly Detection dataset (MVTec AD) \citep{bergmann2021mvtec,bergmann2019mvtec} in Appendix~\ref{sec:additional_mvtec}.

\subsection{Hyperparameter Sensitivity}
\label{sec:hyperparameter}

\begin{figure}
  \centering
    \begin{minipage}{0.245\linewidth}
      \includegraphics[width=\linewidth]{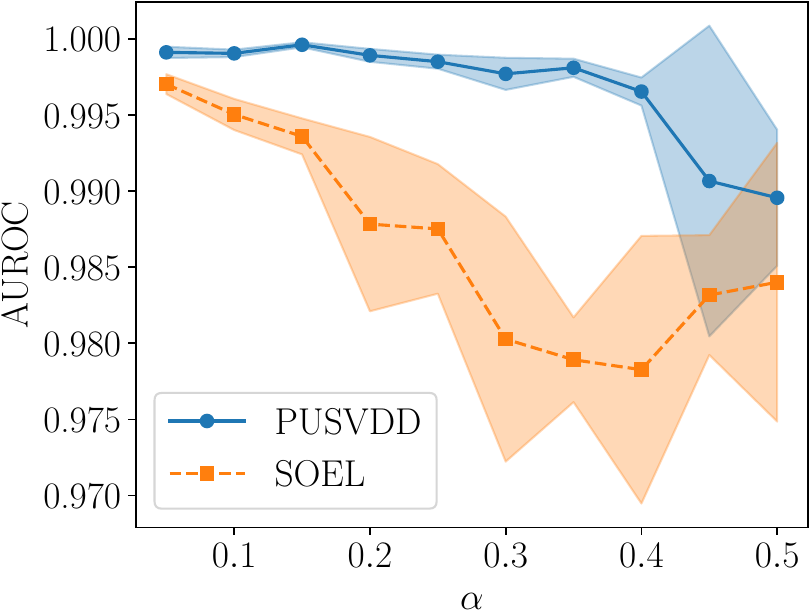}
      \subcaption{MNIST}
      \label{fig:pusvdd_alpha_mnist}
    \end{minipage}
    \begin{minipage}{0.245\linewidth}
      \includegraphics[width=\linewidth]{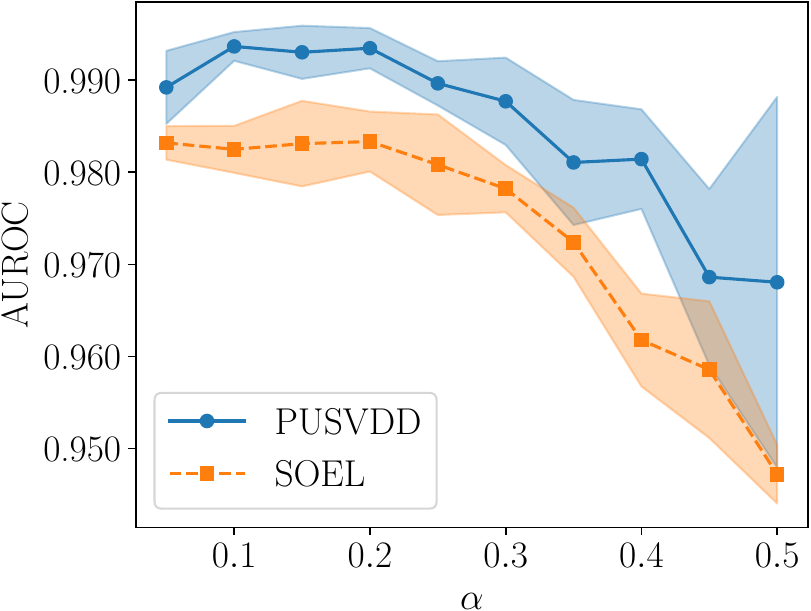}
      \subcaption{FashionMNIST}
      \label{fig:pusvdd_alpha_fashion_mnist}
    \end{minipage}
    \begin{minipage}{0.245\linewidth}
      \includegraphics[width=\linewidth]{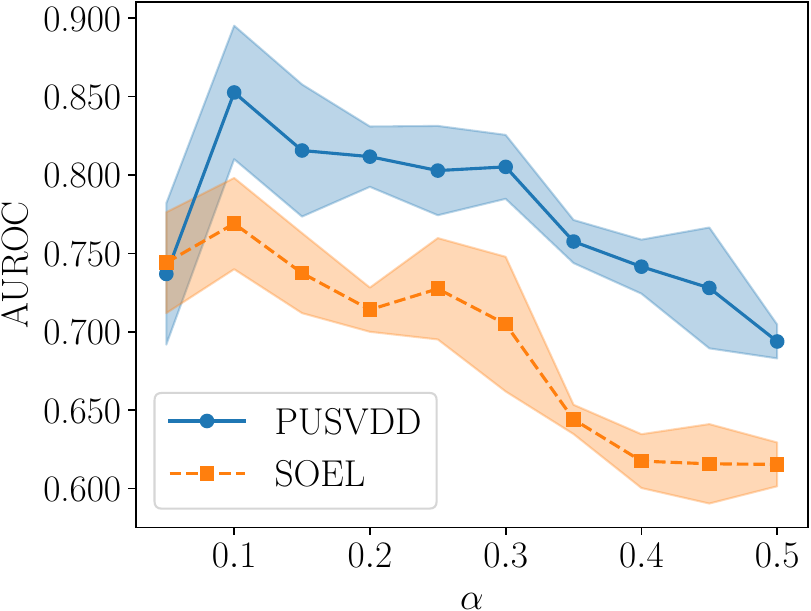}
      \subcaption{SVHN}
      \label{fig:pusvdd_alpha_svhn}
    \end{minipage}
    \begin{minipage}{0.245\linewidth}
      \includegraphics[width=\linewidth]{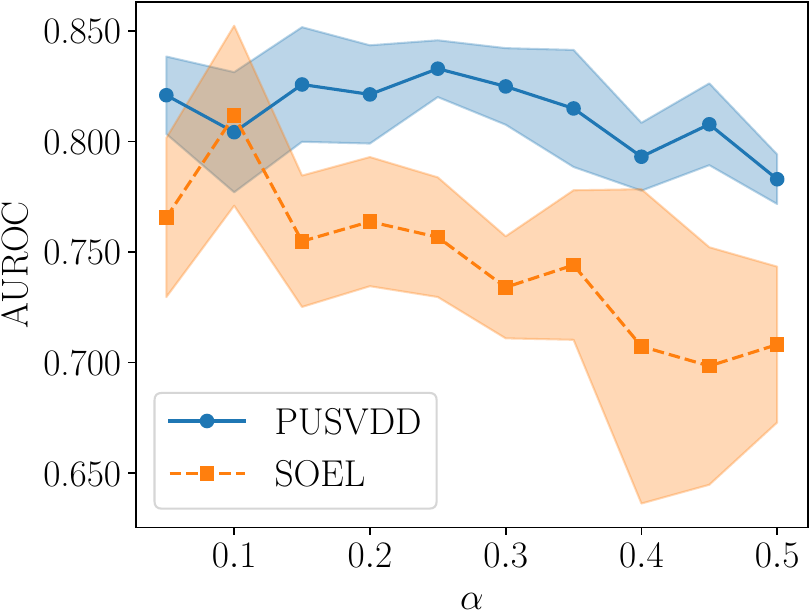}
      \subcaption{CIFAR10}
      \label{fig:pusvdd_alpha_cifar10}
    \end{minipage}
  \caption{
    Relationship between the anomaly detection performance and the hyperparameter $\alpha$ of the PUSVDD and the SOEL on each dataset.
    We used class 0 as unseen anomaly, class 1 as normal, and the remaining classes as seen anomalies.
    The semi-transparent area represents standard deviations.
  }
  \label{fig:pusvdd_alpha}
\end{figure}

Our approach and the SOEL have the hyperparameter $\alpha$, which represents the probability of anomaly occurrence.
In the above experiments, we set it to $\alpha=0.1$ since the 10\% of the training dataset is anomalies.
Finally, we evaluate the sensitivity of $\alpha$.
Figure~\ref{fig:pusvdd_alpha} shows the relationship between the detection performance and the hyperparameter $\alpha$ of the PUSVDD and the SOEL on each dataset.

In most datasets, the PUSVDD and the SOEL achieve the best performance around $\alpha=0.1$.
This indicates that, as with conventional PU learning,
it is important to set $\alpha$ accurately.
Note that $\alpha$ can be estimated from the unlabeled and anomaly training data using conventional PU learning approaches \citep{menon2015learning,ramaswamy2016mixture,jain2016estimating,christoffel2016class}.

Compared to the SOEL, the PUSVDD is more robust to variations in $\alpha$.
In other words, even if $\alpha$ deviates from the true value,
the PUSVDD maintains relatively stable performance.
This is because $\alpha$ in the SOEL is closely related to the number of anomalies within the unlabeled data.
If $\alpha$ deviates from the true value,
normal data may be incorrectly treated as anomalies, or vice versa.
On the other hand,
since $\alpha$ in the PUSVDD only adjusts the weight of the loss function,
it is expected to be relatively robust to deviations in $\alpha$.

Appendix~\ref{sec:alpha_estimation} presents end-to-end experiments across four datasets (MNIST, FashionMNIST, SVHN, and CIFAR-10) in which $\alpha$ is automatically estimated using Class Prior Estimation \citep{christoffel2016class}.
Although CPE estimates deviate substantially from the true $\alpha$ on complex datasets such as SVHN and CIFAR-10, PUSVDD continues to outperform SOEL in all cases, owing to its robustness to deviations in $\alpha$.

\section{Conclusion and Limitations}
\label{sec:conclusion}

Although most unlabeled data are assumed to be normal in semi-supervised anomaly detection,
they are often contaminated with anomalies in practice,
which degrades the detection performance.
To solve this,
we propose the deep positive-unlabeled anomaly detection framework,
which integrates PU learning with deep anomaly detection models such as the AE and the DeepSVDD.
Our approach enables us to approximate the anomaly scores for normal data with the unlabeled data and labeled anomaly data.
Therefore, without the labeled normal data,
we can train the anomaly detector to minimize the anomaly scores for normal data,
and to maximize those for the anomaly data.
Our approach achieves better detection performance than existing approaches on various image datasets.

A limitation of our approach lies in the assumption that the unlabeled data do not contain unseen anomalies.
This assumption follows the same setting as SOEL \citep{li2023deep}, ensuring a fair comparison.
We empirically evaluate the effect of unseen anomaly contamination in the unlabeled data on the toy dataset in Appendix~\ref{sec:unseen_contamination}, while its theoretical analysis is an important direction for future work.
Intuitively, when unseen anomalies constitute a small fraction of the unlabeled data, they are unlikely to substantially affect training, as the base anomaly detector is designed to model the normal class and will naturally assign higher anomaly scores to them.
Conversely, when unseen anomalies are prevalent, they become difficult to distinguish from normal data, potentially degrading detection performance, which is a limitation shared by any method that relies on unlabeled data.
To mitigate this, robust anomaly detectors such as the LOE can be employed as the base detector for our framework.

Another practical consideration concerns the contamination ratio $\alpha$, which is treated as a known hyperparameter in this paper but is typically unavailable in practice.
It can be estimated from data using class prior estimation methods \citep{menon2015learning,ramaswamy2016mixture,jain2016estimating,christoffel2016class}, and Appendix~\ref{sec:alpha_estimation} confirms competitive performance even with an estimated $\hat{\alpha}$.
As shown in Figure~\ref{fig:pusvdd_alpha}, the proposed method is also more robust to deviations in $\alpha$ than SOEL, since $\alpha$ here only adjusts loss weights rather than directly controlling the anomaly sample count.

In addition, extending the framework to non-image domains such as tabular and time-series data is also an important direction for future work.

\section*{Publication status}
Accepted for publication in Neurocomputing:
\url{https://doi.org/10.1016/j.neucom.2026.135134}.

\FloatBarrier
\bibliographystyle{plainnat}
\bibliography{main}

\begin{thebibliography}{71}
\providecommand{\natexlab}[1]{#1}
\providecommand{\url}[1]{\texttt{#1}}
\expandafter\ifx\csname urlstyle\endcsname\relax
  \providecommand{\doi}[1]{doi: #1}\else
  \providecommand{\doi}{doi: \begingroup \urlstyle{rm}\Url}\fi

\bibitem[Bekker and Davis(2020)]{bekker2020learning}
Jessa Bekker and Jesse Davis.
\newblock Learning from positive and unlabeled data: A survey.
\newblock \emph{Machine Learning}, 109:\penalty0 719--760, 2020.

\bibitem[Bergmann et~al.(2019)Bergmann, Fauser, Sattlegger, and
  Steger]{bergmann2019mvtec}
Paul Bergmann, Michael Fauser, David Sattlegger, and Carsten Steger.
\newblock Mvtec ad--a comprehensive real-world dataset for unsupervised anomaly
  detection.
\newblock In \emph{Proceedings of the IEEE/CVF Conference on Computer Vision
  and Pattern Recognition}, pages 9592--9600, 2019.

\bibitem[Bergmann et~al.(2021)Bergmann, Batzner, Fauser, Sattlegger, and
  Steger]{bergmann2021mvtec}
Paul Bergmann, Kilian Batzner, Michael Fauser, David Sattlegger, and Carsten
  Steger.
\newblock The mvtec anomaly detection dataset: a comprehensive real-world
  dataset for unsupervised anomaly detection.
\newblock \emph{International Journal of Computer Vision}, 129\penalty0
  (4):\penalty0 1038--1059, 2021.

\bibitem[Borghesi et~al.(2019)Borghesi, Bartolini, Lombardi, Milano, and
  Benini]{borghesi2019anomaly}
Andrea Borghesi, Andrea Bartolini, Michele Lombardi, Michela Milano, and Luca
  Benini.
\newblock Anomaly detection using autoencoders in high performance computing
  systems.
\newblock In \emph{Proceedings of the AAAI Conference on Artificial
  Intelligence}, volume~33, pages 9428--9433, 2019.

\bibitem[Cerri et~al.(2019)Cerri, Nguyen, Pierini, Spiropulu, and
  Vlimant]{cerri2019variational}
Olmo Cerri, Thong~Q Nguyen, Maurizio Pierini, Maria Spiropulu, and Jean-Roch
  Vlimant.
\newblock Variational autoencoders for new physics mining at the large hadron
  collider.
\newblock \emph{Journal of High Energy Physics}, 2019\penalty0 (5):\penalty0
  1--29, 2019.

\bibitem[Choi et~al.(2018)Choi, Jang, and Alemi]{choi2018waic}
Hyunsun Choi, Eric Jang, and Alexander~A Alemi.
\newblock Waic, but why? generative ensembles for robust anomaly detection.
\newblock \emph{arXiv preprint arXiv:1810.01392}, 2018.

\bibitem[Christoffel et~al.(2016)Christoffel, Niu, and
  Sugiyama]{christoffel2016class}
Marthinus Christoffel, Gang Niu, and Masashi Sugiyama.
\newblock Class-prior estimation for learning from positive and unlabeled data.
\newblock In \emph{Asian Conference on Machine Learning}, pages 221--236. PMLR,
  2016.

\bibitem[Ding et~al.(2022)Ding, Pang, and Shen]{ding2022catching}
Choubo Ding, Guansong Pang, and Chunhua Shen.
\newblock Catching both gray and black swans: Open-set supervised anomaly
  detection.
\newblock In \emph{Proceedings of the IEEE/CVF Conference on Computer Vision
  and Pattern Recognition}, pages 7388--7398, 2022.

\bibitem[Du~Plessis et~al.(2015)Du~Plessis, Niu, and Sugiyama]{du2015convex}
Marthinus Du~Plessis, Gang Niu, and Masashi Sugiyama.
\newblock Convex formulation for learning from positive and unlabeled data.
\newblock In \emph{International Conference on Machine Learning}, pages
  1386--1394. PMLR, 2015.

\bibitem[Du~Plessis et~al.(2014)Du~Plessis, Niu, and Sugiyama]{du2014analysis}
Marthinus~C Du~Plessis, Gang Niu, and Masashi Sugiyama.
\newblock Analysis of learning from positive and unlabeled data.
\newblock \emph{Advances in Neural Information Processing Systems}, 27, 2014.

\bibitem[Durani et~al.(2025)Durani, Nitzl, Plant, and
  B{\"o}hm]{durani2025weakly}
Walid Durani, Tobias Nitzl, Claudia Plant, and Christian B{\"o}hm.
\newblock Weakly supervised anomaly detection via dual-tailed kernel.
\newblock In \emph{International Conference on Machine Learning}, 2025.

\bibitem[Elkan and Noto(2008)]{elkan2008learning}
Charles Elkan and Keith Noto.
\newblock Learning classifiers from only positive and unlabeled data.
\newblock In \emph{Proceedings of the ACM SIGKDD Conference on Knowledge
  Discovery and Data Mining}, pages 213--220, 2008.

\bibitem[Goodfellow et~al.(2014)Goodfellow, Pouget-Abadie, Mirza, Xu,
  Warde-Farley, Ozair, Courville, and Bengio]{goodfellow2014generative}
Ian Goodfellow, Jean Pouget-Abadie, Mehdi Mirza, Bing Xu, David Warde-Farley,
  Sherjil Ozair, Aaron Courville, and Yoshua Bengio.
\newblock Generative adversarial nets.
\newblock \emph{Advances in Neural Information Processing Systems}, 27, 2014.

\bibitem[Goodfellow et~al.(2016)Goodfellow, Bengio, and
  Courville]{goodfellow2016deep}
Ian Goodfellow, Yoshua Bengio, and Aaron Courville.
\newblock \emph{Deep learning}.
\newblock MIT press, 2016.

\bibitem[Hammoudeh and Lowd(2020)]{hammoudeh2020learning}
Zayd Hammoudeh and Daniel Lowd.
\newblock Learning from positive and unlabeled data with arbitrary positive
  shift.
\newblock \emph{Advances in Neural Information Processing Systems},
  33:\penalty0 13088--13099, 2020.

\bibitem[Havtorn et~al.(2021)Havtorn, Frellsen, Hauberg, and
  Maal{\o}e]{havtorn2021hierarchical}
Jakob~D Havtorn, Jes Frellsen, S{\o}ren Hauberg, and Lars Maal{\o}e.
\newblock Hierarchical vaes know what they don’t know.
\newblock In \emph{International Conference on Machine Learning}, pages
  4117--4128. PMLR, 2021.

\bibitem[He et~al.(2016)He, Zhang, Ren, and Sun]{he2016deep}
Kaiming He, Xiangyu Zhang, Shaoqing Ren, and Jian Sun.
\newblock Deep residual learning for image recognition.
\newblock In \emph{Proceedings of the IEEE/CVF Conference on Computer Vision
  and Pattern Recognition}, pages 770--778, 2016.

\bibitem[Hendrycks et~al.(2018)Hendrycks, Mazeika, and
  Dietterich]{hendrycks2018deep}
Dan Hendrycks, Mantas Mazeika, and Thomas Dietterich.
\newblock Deep anomaly detection with outlier exposure.
\newblock In \emph{International Conference on Learning Representations}, 2018.

\bibitem[Hinton and Salakhutdinov(2006)]{hinton2006reducing}
Geoffrey~E Hinton and Ruslan~R Salakhutdinov.
\newblock Reducing the dimensionality of data with neural networks.
\newblock \emph{science}, 313\penalty0 (5786):\penalty0 504--507, 2006.

\bibitem[Jain et~al.(2016)Jain, White, and Radivojac]{jain2016estimating}
Shantanu Jain, Martha White, and Predrag Radivojac.
\newblock Estimating the class prior and posterior from noisy positives and
  unlabeled data.
\newblock \emph{Advances in Neural Information Processing Systems}, 29, 2016.

\bibitem[Ju et~al.(2020)Ju, Lee, Hwang, Namkung, and Yu]{ju2020pumad}
Hyunjun Ju, Dongha Lee, Junyoung Hwang, Junghyun Namkung, and Hwanjo Yu.
\newblock Pumad: Pu metric learning for anomaly detection.
\newblock \emph{Information Sciences}, 523:\penalty0 167--183, 2020.

\bibitem[Kingma and Ba(2015)]{kingma2015adam}
Diederik~P. Kingma and Jimmy Ba.
\newblock Adam: A method for stochastic optimization.
\newblock In \emph{International Conference on Learning Representations}, 2015.

\bibitem[Kingma and Welling(2014)]{kingma2014auto}
Diederik~P Kingma and Max Welling.
\newblock Auto-encoding variational {B}ayes.
\newblock In \emph{International Conference on Learning Representations}, 2014.

\bibitem[Kingma et~al.(2015)Kingma, Salimans, and
  Welling]{kingma2015variational}
Durk~P Kingma, Tim Salimans, and Max Welling.
\newblock Variational dropout and the local reparameterization trick.
\newblock \emph{Advances in Neural Information Processing Systems}, 28, 2015.

\bibitem[Kiryo et~al.(2017)Kiryo, Niu, Du~Plessis, and
  Sugiyama]{kiryo2017positive}
Ryuichi Kiryo, Gang Niu, Marthinus~C Du~Plessis, and Masashi Sugiyama.
\newblock Positive-unlabeled learning with non-negative risk estimator.
\newblock \emph{Advances in Neural Information Processing Systems}, 30, 2017.

\bibitem[Krizhevsky et~al.(2009)Krizhevsky, Hinton,
  et~al.]{krizhevsky2009learning}
Alex Krizhevsky, Geoffrey Hinton, et~al.
\newblock Learning multiple layers of features from tiny images.
\newblock 2009.

\bibitem[Kumagai et~al.(2025)Kumagai, Iwata, Takahashi, Nishiyama, and
  Fujiwara]{kumagai2025importance}
Atsutoshi Kumagai, Tomoharu Iwata, Hiroshi Takahashi, Taishi Nishiyama, and
  Yasuhiro Fujiwara.
\newblock Importance-weighted positive-unlabeled learning for distribution
  shift adaptation.
\newblock In \emph{International Conference on Artificial Intelligence and
  Statistics}, pages 1576--1584. PMLR, 2025.

\bibitem[Kwon et~al.(2019)Kwon, Kim, Kim, Suh, Kim, and Kim]{kwon2019survey}
Donghwoon Kwon, Hyunjoo Kim, Jinoh Kim, Sang~C Suh, Ikkyun Kim, and Kuinam~J
  Kim.
\newblock A survey of deep learning-based network anomaly detection.
\newblock \emph{Cluster Computing}, 22:\penalty0 949--961, 2019.

\bibitem[Li et~al.(2023)Li, Qiu, Kloft, Smyth, Mandt, and Rudolph]{li2023deep}
Aodong Li, Chen Qiu, Marius Kloft, Padhraic Smyth, Stephan Mandt, and Maja
  Rudolph.
\newblock Deep anomaly detection under labeling budget constraints.
\newblock In \emph{International Conference on Machine Learning}, pages
  19882--19910. PMLR, 2023.

\bibitem[Litjens et~al.(2017)Litjens, Kooi, Bejnordi, Setio, Ciompi,
  Ghafoorian, Van Der~Laak, Van~Ginneken, and S{\'a}nchez]{litjens2017survey}
Geert Litjens, Thijs Kooi, Babak~Ehteshami Bejnordi, Arnaud Arindra~Adiyoso
  Setio, Francesco Ciompi, Mohsen Ghafoorian, Jeroen~Awm Van Der~Laak, Bram
  Van~Ginneken, and Clara~I S{\'a}nchez.
\newblock A survey on deep learning in medical image analysis.
\newblock \emph{Medical image analysis}, 42:\penalty0 60--88, 2017.

\bibitem[Liu et~al.(2008)Liu, Ting, and Zhou]{liu2008isolation}
Fei~Tony Liu, Kai~Ming Ting, and Zhi-Hua Zhou.
\newblock Isolation forest.
\newblock In \emph{2008 eighth ieee international conference on data mining},
  pages 413--422. IEEE, 2008.

\bibitem[Marchi et~al.(2015)Marchi, Vesperini, Eyben, Squartini, and
  Schuller]{marchi2015novel}
Erik Marchi, Fabio Vesperini, Florian Eyben, Stefano Squartini, and Bj{\"o}rn
  Schuller.
\newblock A novel approach for automatic acoustic novelty detection using a
  denoising autoencoder with bidirectional lstm neural networks.
\newblock In \emph{2015 IEEE international conference on acoustics, speech and
  signal processing (ICASSP)}, pages 1996--2000. IEEE, 2015.

\bibitem[Menon et~al.(2015)Menon, Van~Rooyen, Ong, and
  Williamson]{menon2015learning}
Aditya Menon, Brendan Van~Rooyen, Cheng~Soon Ong, and Bob Williamson.
\newblock Learning from corrupted binary labels via class-probability
  estimation.
\newblock In \emph{International Conference on Machine Learning}, pages
  125--134. PMLR, 2015.

\bibitem[Min et~al.(2017)Min, Lee, and Yoon]{min2017deep}
Seonwoo Min, Byunghan Lee, and Sungroh Yoon.
\newblock Deep learning in bioinformatics.
\newblock \emph{Briefings in bioinformatics}, 18\penalty0 (5):\penalty0
  851--869, 2017.

\bibitem[Mohri et~al.(2018)Mohri, Rostamizadeh, and
  Talwalkar]{mohri2018foundations}
Mehryar Mohri, Afshin Rostamizadeh, and Ameet Talwalkar.
\newblock \emph{Foundations of machine learning}.
\newblock MIT press, 2018.

\bibitem[Nakajima and Sugiyama(2023)]{nakajima2023positive}
Shota Nakajima and Masashi Sugiyama.
\newblock Positive-unlabeled classification under class-prior shift: a
  prior-invariant approach based on density ratio estimation.
\newblock \emph{Machine Learning}, 112\penalty0 (3):\penalty0 889--919, 2023.

\bibitem[Nalisnick et~al.(2018)Nalisnick, Matsukawa, Teh, Gorur, and
  Lakshminarayanan]{nalisnick2018deep}
Eric Nalisnick, Akihiro Matsukawa, Yee~Whye Teh, Dilan Gorur, and Balaji
  Lakshminarayanan.
\newblock Do deep generative models know what they don't know?
\newblock \emph{arXiv preprint arXiv:1810.09136}, 2018.

\bibitem[Natarajan et~al.(2013)Natarajan, Dhillon, Ravikumar, and
  Tewari]{natarajan2013learning}
Nagarajan Natarajan, Inderjit~S Dhillon, Pradeep~K Ravikumar, and Ambuj Tewari.
\newblock Learning with noisy labels.
\newblock \emph{Advances in Neural Information Processing Systems}, 26, 2013.

\bibitem[Netzer et~al.(2011)Netzer, Wang, Coates, Bissacco, Wu, and
  Ng]{netzer2011reading}
Yuval Netzer, Tao Wang, Adam Coates, Alessandro Bissacco, Bo~Wu, and Andrew~Y
  Ng.
\newblock Reading digits in natural images with unsupervised feature learning.
\newblock In \emph{NIPS Workshop on Deep Learning and Unsupervised Feature
  Learning 2011}, 2011.

\bibitem[Niu et~al.(2016)Niu, Du~Plessis, Sakai, Ma, and
  Sugiyama]{niu2016theoretical}
Gang Niu, Marthinus~Christoffel Du~Plessis, Tomoya Sakai, Yao Ma, and Masashi
  Sugiyama.
\newblock Theoretical comparisons of positive-unlabeled learning against
  positive-negative learning.
\newblock \emph{Advances in Neural Information Processing Systems}, 29, 2016.

\bibitem[Pang et~al.(2021)Pang, van~den Hengel, Shen, and Cao]{pang2021toward}
Guansong Pang, Anton van~den Hengel, Chunhua Shen, and Longbing Cao.
\newblock Toward deep supervised anomaly detection: Reinforcement learning from
  partially labeled anomaly data.
\newblock In \emph{Proceedings of the ACM SIGKDD Conference on Knowledge
  Discovery and Data Mining}, pages 1298--1308, 2021.

\bibitem[Pang et~al.(2023)Pang, Shen, Jin, and van~den Hengel]{pang2023deep}
Guansong Pang, Chunhua Shen, Huidong Jin, and Anton van~den Hengel.
\newblock Deep weakly-supervised anomaly detection.
\newblock In \emph{Proceedings of the ACM SIGKDD Conference on Knowledge
  Discovery and Data Mining}, pages 1795--1807, 2023.

\bibitem[Patrini et~al.(2016)Patrini, Nielsen, Nock, and
  Carioni]{patrini2016loss}
Giorgio Patrini, Frank Nielsen, Richard Nock, and Marcello Carioni.
\newblock Loss factorization, weakly supervised learning and label noise
  robustness.
\newblock In \emph{International Conference on Machine Learning}, pages
  708--717. PMLR, 2016.

\bibitem[Pedregosa et~al.(2011)Pedregosa, Varoquaux, Gramfort, Michel, Thirion,
  Grisel, Blondel, Prettenhofer, Weiss, Dubourg, et~al.]{pedregosa2011scikit}
Fabian Pedregosa, Ga{\"e}l Varoquaux, Alexandre Gramfort, Vincent Michel,
  Bertrand Thirion, Olivier Grisel, Mathieu Blondel, Peter Prettenhofer, Ron
  Weiss, Vincent Dubourg, et~al.
\newblock Scikit-learn: Machine learning in python.
\newblock \emph{the Journal of machine Learning research}, 12:\penalty0
  2825--2830, 2011.

\bibitem[Perera et~al.(2019)Perera, Nallapati, and Xiang]{perera2019ocgan}
Pramuditha Perera, Ramesh Nallapati, and Bing Xiang.
\newblock Ocgan: One-class novelty detection using gans with constrained latent
  representations.
\newblock In \emph{Proceedings of the IEEE/CVF Conference on Computer Vision
  and Pattern Recognition}, pages 2898--2906, 2019.

\bibitem[Perini et~al.(2023)Perini, Vercruyssen, and Davis]{perini2023learning}
Lorenzo Perini, Vincent Vercruyssen, and Jesse Davis.
\newblock Learning from positive and unlabeled multi-instance bags in anomaly
  detection.
\newblock In \emph{Proceedings of the ACM SIGKDD Conference on Knowledge
  Discovery and Data Mining}, pages 1897--1906, 2023.

\bibitem[Pracht et~al.(2020)Pracht, Bohle, and Grimme]{pracht2020automated}
Philipp Pracht, Fabian Bohle, and Stefan Grimme.
\newblock Automated exploration of the low-energy chemical space with fast
  quantum chemical methods.
\newblock \emph{Physical Chemistry Chemical Physics}, 22\penalty0
  (14):\penalty0 7169--7192, 2020.

\bibitem[Qiu et~al.(2022)Qiu, Li, Kloft, Rudolph, and Mandt]{qiu2022latent}
Chen Qiu, Aodong Li, Marius Kloft, Maja Rudolph, and Stephan Mandt.
\newblock Latent outlier exposure for anomaly detection with contaminated data.
\newblock In \emph{International Conference on Machine Learning}, pages
  18153--18167. PMLR, 2022.

\bibitem[Ramaswamy et~al.(2016)Ramaswamy, Scott, and
  Tewari]{ramaswamy2016mixture}
Harish Ramaswamy, Clayton Scott, and Ambuj Tewari.
\newblock Mixture proportion estimation via kernel embeddings of distributions.
\newblock In \emph{International Conference on Machine Learning}, pages
  2052--2060. PMLR, 2016.

\bibitem[Ren et~al.(2019)Ren, Liu, Fertig, Snoek, Poplin, Depristo, Dillon, and
  Lakshminarayanan]{ren2019likelihood}
Jie Ren, Peter~J Liu, Emily Fertig, Jasper Snoek, Ryan Poplin, Mark Depristo,
  Joshua Dillon, and Balaji Lakshminarayanan.
\newblock Likelihood ratios for out-of-distribution detection.
\newblock \emph{Advances in Neural Information Processing Systems}, 32, 2019.

\bibitem[Ruff et~al.(2018)Ruff, Vandermeulen, Goernitz, Deecke, Siddiqui,
  Binder, M{\"u}ller, and Kloft]{ruff2018deep}
Lukas Ruff, Robert Vandermeulen, Nico Goernitz, Lucas Deecke, Shoaib~Ahmed
  Siddiqui, Alexander Binder, Emmanuel M{\"u}ller, and Marius Kloft.
\newblock Deep one-class classification.
\newblock In \emph{International Conference on Machine Learning}, pages
  4393--4402. PMLR, 2018.

\bibitem[Ruff et~al.(2019)Ruff, Vandermeulen, G{\"o}rnitz, Binder, M{\"u}ller,
  M{\"u}ller, and Kloft]{ruff2019deep}
Lukas Ruff, Robert~A Vandermeulen, Nico G{\"o}rnitz, Alexander Binder, Emmanuel
  M{\"u}ller, Klaus-Robert M{\"u}ller, and Marius Kloft.
\newblock Deep semi-supervised anomaly detection.
\newblock In \emph{International Conference on Learning Representations}, 2019.

\bibitem[Ruff et~al.(2021)Ruff, Kauffmann, Vandermeulen, Montavon, Samek,
  Kloft, Dietterich, and M{\"u}ller]{ruff2021unifying}
Lukas Ruff, Jacob~R Kauffmann, Robert~A Vandermeulen, Gr{\'e}goire Montavon,
  Wojciech Samek, Marius Kloft, Thomas~G Dietterich, and Klaus-Robert
  M{\"u}ller.
\newblock A unifying review of deep and shallow anomaly detection.
\newblock \emph{Proceedings of the IEEE}, 109\penalty0 (5):\penalty0 756--795,
  2021.

\bibitem[Sakurada and Yairi(2014)]{sakurada2014anomaly}
Mayu Sakurada and Takehisa Yairi.
\newblock Anomaly detection using autoencoders with nonlinear dimensionality
  reduction.
\newblock In \emph{Proceedings of the MLSDA 2014 2nd workshop on machine
  learning for sensory data analysis}, pages 4--11, 2014.

\bibitem[Salakhutdinov and Murray(2008)]{salakhutdinov2008quantitative}
Ruslan Salakhutdinov and Iain Murray.
\newblock On the quantitative analysis of deep belief networks.
\newblock In \emph{Proceedings of the 25th international conference on Machine
  learning}, pages 872--879. ACM, 2008.

\bibitem[Serr{\`a} et~al.(2019)Serr{\`a}, {\'A}lvarez, G{\'o}mez, Slizovskaia,
  N{\'u}{\~n}ez, and Luque]{serra2019input}
Joan Serr{\`a}, David {\'A}lvarez, Vicen{\c{c}} G{\'o}mez, Olga Slizovskaia,
  Jos{\'e}~F N{\'u}{\~n}ez, and Jordi Luque.
\newblock Input complexity and out-of-distribution detection with
  likelihood-based generative models.
\newblock In \emph{International Conference on Learning Representations}, 2019.

\bibitem[Shang et~al.(2023)Shang, Zhao, Yan, and Chen]{shang2023core}
Zuogang Shang, Zhibin Zhao, Ruqiang Yan, and Xuefeng Chen.
\newblock Core loss: Mining core samples efficiently for robust machine anomaly
  detection against data pollution.
\newblock \emph{Mechanical Systems and Signal Processing}, 189:\penalty0
  110046, 2023.

\bibitem[Shou et~al.(2025)Shou, Lu, Pavlovski, and Zhou]{shou2025read}
Hongzhe Shou, Guanyu Lu, Martin Pavlovski, and Fang Zhou.
\newblock Read: Robust and efficient anomaly detection under data contamination
  and limited supervision.
\newblock In \emph{Proceedings of the ACM SIGKDD Conference on Knowledge
  Discovery and Data Mining}, pages 2586--2596, 2025.

\bibitem[Tax and Duin(2004)]{tax2004support}
David~MJ Tax and Robert~PW Duin.
\newblock Support vector data description.
\newblock \emph{Machine learning}, 54:\penalty0 45--66, 2004.

\bibitem[Vincent et~al.(2008)Vincent, Larochelle, Bengio, and
  Manzagol]{vincent2008extracting}
Pascal Vincent, Hugo Larochelle, Yoshua Bengio, and Pierre-Antoine Manzagol.
\newblock Extracting and composing robust features with denoising autoencoders.
\newblock In \emph{Proceedings of the 25th international conference on Machine
  learning}, pages 1096--1103, 2008.

\bibitem[Wang et~al.(2013)Wang, Jajodia, Singhal, Cheng, and Noel]{wang2013k}
Lingyu Wang, Sushil Jajodia, Anoop Singhal, Pengsu Cheng, and Steven Noel.
\newblock k-zero day safety: A network security metric for measuring the risk
  of unknown vulnerabilities.
\newblock \emph{IEEE Transactions on Dependable and Secure Computing},
  11\penalty0 (1):\penalty0 30--44, 2013.

\bibitem[Xiao et~al.(2017)Xiao, Rasul, and Vollgraf]{xiao2017fashion}
Han Xiao, Kashif Rasul, and Roland Vollgraf.
\newblock Fashion-{MNIST}: a novel image dataset for benchmarking machine
  learning algorithms.
\newblock \emph{arXiv preprint arXiv:1708.07747}, 2017.

\bibitem[Xiao et~al.(2020)Xiao, Yan, and Amit]{xiao2020likelihood}
Zhisheng Xiao, Qing Yan, and Yali Amit.
\newblock Likelihood regret: An out-of-distribution detection score for
  variational auto-encoder.
\newblock \emph{Advances in Neural Information Processing Systems},
  33:\penalty0 20685--20696, 2020.

\bibitem[Yamanaka et~al.(2019)Yamanaka, Iwata, Takahashi, Yamada, and
  Kanai]{yamanaka2019autoencoding}
Yuki Yamanaka, Tomoharu Iwata, Hiroshi Takahashi, Masanori Yamada, and
  Sekitoshi Kanai.
\newblock Autoencoding binary classifiers for supervised anomaly detection.
\newblock In \emph{PRICAI 2019: Trends in Artificial Intelligence: 16th Pacific
  Rim International Conference on Artificial Intelligence, Cuvu, Yanuca Island,
  Fiji, August 26--30, 2019, Proceedings, Part II 16}, pages 647--659.
  Springer, 2019.

\bibitem[Yang et~al.(2021)Yang, Shi, and Ni]{yang2021medmnist}
Jiancheng Yang, Rui Shi, and Bingbing Ni.
\newblock Medmnist classification decathlon: A lightweight automl benchmark for
  medical image analysis.
\newblock In \emph{2021 IEEE 18th International Symposium on Biomedical Imaging
  (ISBI)}, pages 191--195. IEEE, 2021.

\bibitem[Yang et~al.(2023)Yang, Shi, Wei, Liu, Zhao, Ke, Pfister, and
  Ni]{yang2023medmnist}
Jiancheng Yang, Rui Shi, Donglai Wei, Zequan Liu, Lin Zhao, Bilian Ke,
  Hanspeter Pfister, and Bingbing Ni.
\newblock Medmnist v2-a large-scale lightweight benchmark for 2d and 3d
  biomedical image classification.
\newblock \emph{Scientific Data}, 10\penalty0 (1):\penalty0 41, 2023.

\bibitem[Yoon et~al.(2021)Yoon, Noh, and Park]{yoon2021autoencoding}
Sangwoong Yoon, Yung-Kyun Noh, and Frank Park.
\newblock Autoencoding under normalization constraints.
\newblock In \emph{International Conference on Machine Learning}, pages
  12087--12097. PMLR, 2021.

\bibitem[Zhang et~al.(2021)Zhang, Cao, VanNostrand, Madden, and
  Rundensteiner]{zhang2021elite}
Huayi Zhang, Lei Cao, Peter VanNostrand, Samuel Madden, and Elke~A
  Rundensteiner.
\newblock Elite: Robust deep anomaly detection with meta gradient.
\newblock In \emph{Proceedings of the ACM SIGKDD Conference on Knowledge
  Discovery and Data Mining}, pages 2174--2182, 2021.

\bibitem[Zhang et~al.(2018)Zhang, Wang, Meng, Tan, and Yuan]{zhang2018boosting}
Jiaqi Zhang, Zhenzhen Wang, Jingjing Meng, Yap-Peng Tan, and Junsong Yuan.
\newblock Boosting positive and unlabeled learning for anomaly detection with
  multi-features.
\newblock \emph{IEEE Transactions on Multimedia}, 21\penalty0 (5):\penalty0
  1332--1344, 2018.

\bibitem[Zhou and Paffenroth(2017)]{zhou2017anomaly}
Chong Zhou and Randy~C Paffenroth.
\newblock Anomaly detection with robust deep autoencoders.
\newblock In \emph{Proceedings of the ACM SIGKDD Conference on Knowledge
  Discovery and Data Mining}, pages 665--674, 2017.

\bibitem[Zhou et~al.(2021)Zhou, Song, Zhang, Liu, Zhu, and
  Liu]{zhou2021feature}
Yingjie Zhou, Xucheng Song, Yanru Zhang, Fanxing Liu, Ce~Zhu, and Lingqiao Liu.
\newblock Feature encoding with autoencoders for weakly supervised anomaly
  detection.
\newblock \emph{IEEE Transactions on Neural Networks and Learning Systems},
  33\penalty0 (6):\penalty0 2454--2465, 2021.

\end{thebibliography}

\clearpage
\appendix

\section{Pseudo Code}
\label{sec:pseudo_code}

\begin{algorithm}[H]
  \caption{Positive-Unlabeled Autoencoder}
  \begin{algorithmic}[1]
    \Require Unlabeled and anomaly datasets $(\mathcal{U}, \mathcal{A})$, hyperparameter $\alpha\in[0,1]$
    \Ensure Model parameter $\theta$
    \While{not converged}
      \State Sample mini-batch $\mathcal{B}$ from datasets $(\mathcal{U}, \mathcal{A})$
      \State Compute $\mathcal{L}_{\mathcal{A}}^{+}(\theta)$, $\mathcal{L}_{\mathcal{U}}^{-}(\theta)$, and $\mathcal{L}_{\mathcal{A}}^{-}(\theta)$ in Eq.~(\ref{eq:pn_loss_approx}) with $\mathcal{B}$
      \State Set the gradient $\nabla_{\theta}\left(\alpha\mathcal{L}_{\mathcal{A}}^{+}(\theta)+\left|\mathcal{L}_{\mathcal{U}}^{-}(\theta)-\alpha\mathcal{L}_{\mathcal{A}}^{-}(\theta)\right|\right)$
      \State Update $\theta$ with the gradient
    \EndWhile
  \end{algorithmic}
  \label{alg:proposed}
\end{algorithm}

Algorithm~\ref{alg:proposed} shows the pseudo code of the PUAE.

\section{Dataset Example}
\label{sec:dataset_example}

\begin{figure}[h]
  \centering
    \begin{minipage}{0.245\linewidth}
      \includegraphics[width=\linewidth]{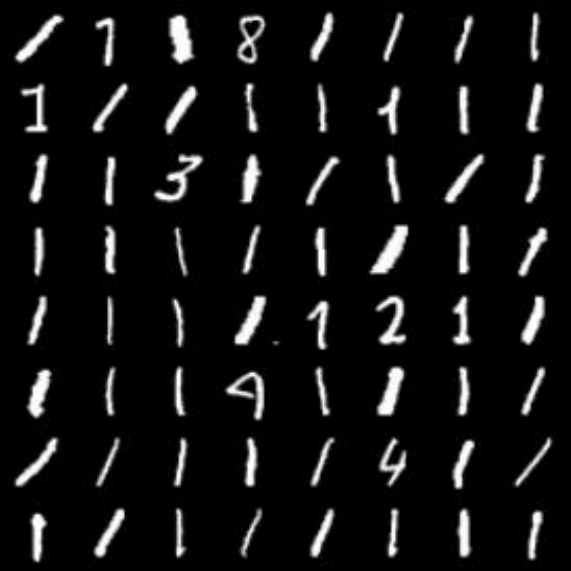}
      \subcaption{Training (unlabeled)}
      \label{fig:train_unlabeled}
    \end{minipage}
    \begin{minipage}{0.245\linewidth}
      \includegraphics[width=\linewidth]{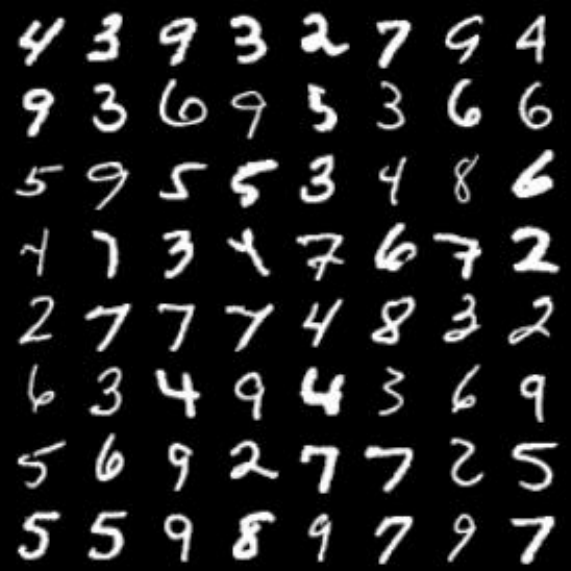}
      \subcaption{Training (anomaly)}
      \label{fig:train_anomaly}
    \end{minipage}
    \begin{minipage}{0.245\linewidth}
      \includegraphics[width=\linewidth]{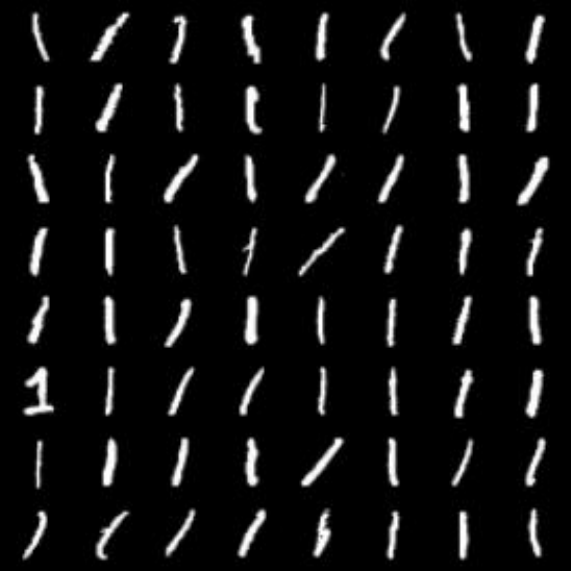}
      \subcaption{Test (normal)}
      \label{fig:test_unlabeled}
    \end{minipage}
    \begin{minipage}{0.245\linewidth}
      \includegraphics[width=\linewidth]{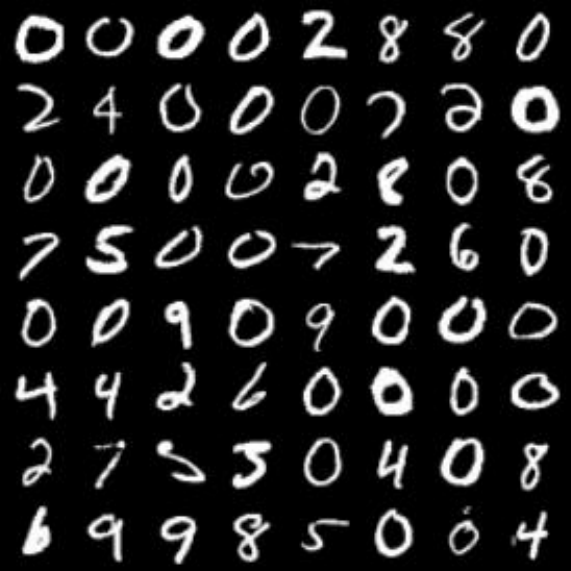}
      \subcaption{Test (anomaly)}
      \label{fig:test_anomaly}
    \end{minipage}
  \caption{
    The example of the dataset in the case of MNIST.
  }
  \label{fig:dataset_example}
\end{figure}

Figure~\ref{fig:dataset_example} shows the example of the dataset in the case of MNIST.
In this example,
the digit 1 is normal,
the digit 0 is unseen anomaly,
and the digits 2, 3, 4, 5, 6, 7, 8, and 9 are seen anomalies.
(a) The unlabeled data points in the training dataset are contaminated with seen anomalies.
(b) The anomaly data points in the training dataset contain seen anomalies but not unseen anomalies.
(c) The normal data points in the test dataset are not contaminated with anomalies.
(d) The anomaly data points in the test dataset contain both seen and unseen anomalies.

\section{Anomaly Detection Performance with Various Numbers of Unlabeled Anomalies}
\label{sec:additional_n_unlabeled_anomalies}

\begin{figure}[h]
  \centering
    \begin{minipage}{0.245\linewidth}
      \includegraphics[width=\linewidth]{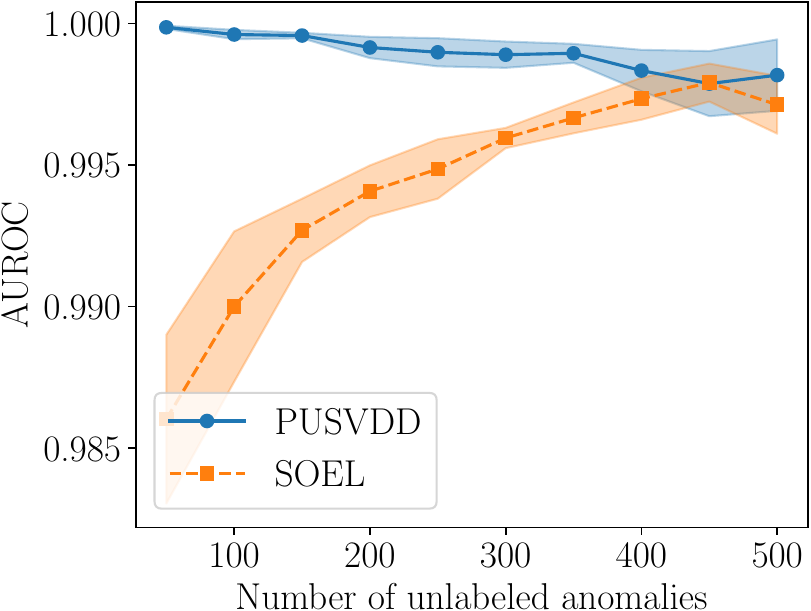}
      \subcaption{MNIST}
      \label{fig:pusvdd_n_unlabeled_anomalies_mnist}
    \end{minipage}
    \begin{minipage}{0.245\linewidth}
      \includegraphics[width=\linewidth]{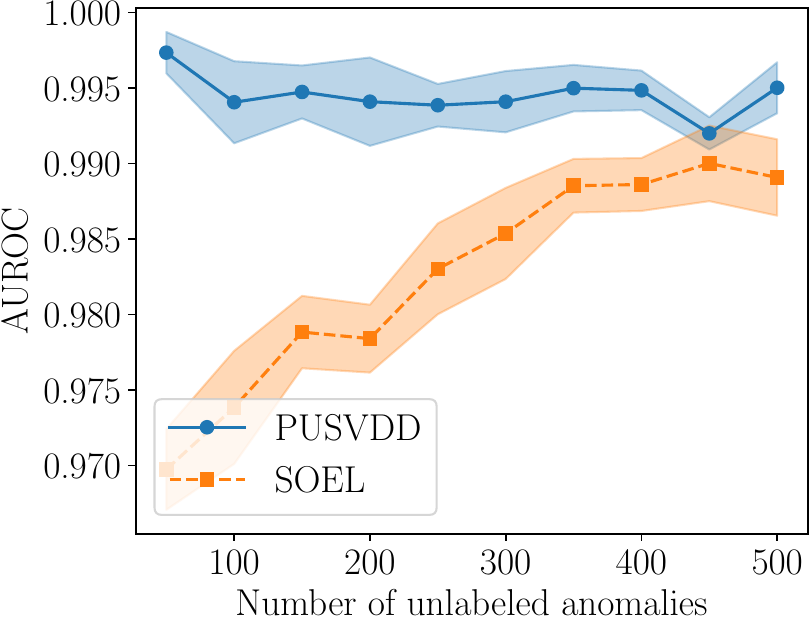}
      \subcaption{FashionMNIST}
      \label{fig:pusvdd_n_unlabeled_anomalies_fashion_mnist}
    \end{minipage}
    \begin{minipage}{0.245\linewidth}
      \includegraphics[width=\linewidth]{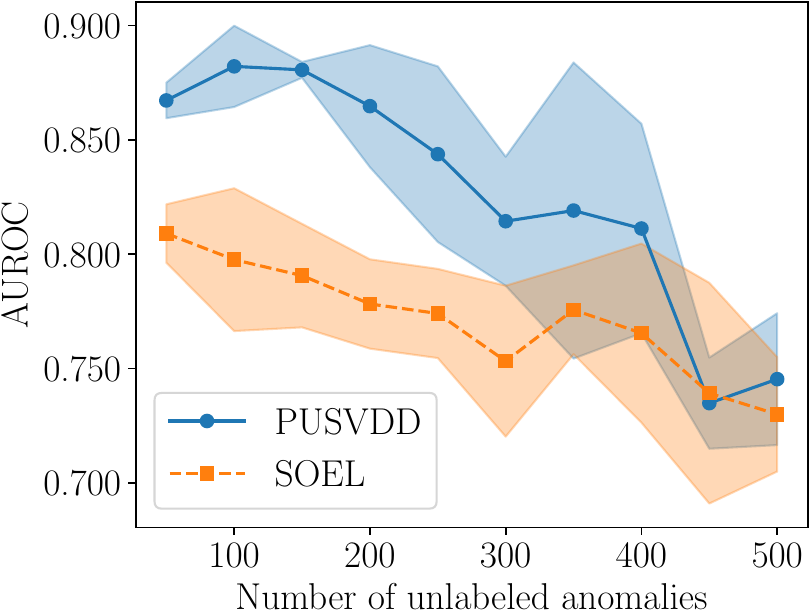}
      \subcaption{SVHN}
      \label{fig:pusvdd_n_unlabeled_anomalies_svhn}
    \end{minipage}
    \begin{minipage}{0.245\linewidth}
      \includegraphics[width=\linewidth]{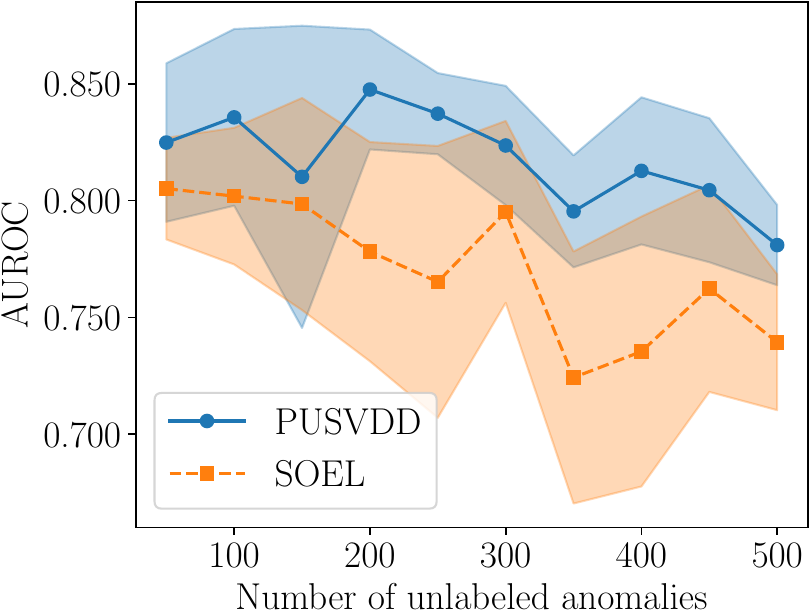}
      \subcaption{CIFAR10}
      \label{fig:pusvdd_n_unlabeled_anomalies_cifar10}
    \end{minipage}
  \caption{
    Comparison of anomaly detection performance between the PUSVDD and the SOEL with various numbers of unlabeled anomalies on each dataset.
    We used class 0 as unseen anomaly, class 1 as normal, and the remaining classes as seen anomalies.
    The semi-transparent area represents standard deviations.
  }
  \label{fig:pusvdd_n_unlabeled_anomalies}
\end{figure}

Figure~\ref{fig:pusvdd_n_unlabeled_anomalies} shows the anomaly detection performance with various numbers of unlabeled anomalies.
As the number of unlabeled anomalies changes, the true contamination rate $\alpha$ of the unlabeled data also changes accordingly;
the hyperparameter $\alpha$ was set to its true value for each case.
Even when the true contamination rate changes, our PUSVDD consistently achieves performance equal to or better than SOEL.

\section{Anomaly Detection Performance for Seen and Unseen Anomalies}
\label{sec:additional_each}

As the additional experiments,
we evaluate the anomaly detection performance for seen and unseen anomalies on each dataset.
The setups are the same as those described in experimental section.

\begin{table}[h]
\caption{
  Seen anomaly detection performance on MNIST, FashionMNIST, SVHN, and CIFAR10.
}
\centering
\begin{tabular}{lrrrr}
\toprule
 & MNIST & FashionMNIST & SVHN & CIFAR10 \\
\midrule
IF & 0.814 $\pm$ 0.096 & 0.915 $\pm$ 0.053 & 0.510 $\pm$ 0.016 & 0.585 $\pm$ 0.106 \\
AE & 0.843 $\pm$ 0.073 & 0.840 $\pm$ 0.092 & 0.563 $\pm$ 0.039 & 0.573 $\pm$ 0.131 \\
DeepSVDD & 0.925 $\pm$ 0.056 & 0.942 $\pm$ 0.040 & 0.600 $\pm$ 0.033 & 0.694 $\pm$ 0.069 \\
LOE & 0.942 $\pm$ 0.038 & 0.940 $\pm$ 0.042 & 0.645 $\pm$ 0.055 & 0.713 $\pm$ 0.077 \\
\midrule
ABC & 0.852 $\pm$ 0.072 & 0.841 $\pm$ 0.092 & 0.564 $\pm$ 0.039 & 0.572 $\pm$ 0.131 \\
DeepSAD & 0.930 $\pm$ 0.052 & 0.956 $\pm$ 0.032 & 0.674 $\pm$ 0.031 & 0.716 $\pm$ 0.074 \\
SOEL & 0.967 $\pm$ 0.024 & 0.963 $\pm$ 0.028 & 0.751 $\pm$ 0.035 & 0.773 $\pm$ 0.073 \\
PU & 0.959 $\pm$ 0.036 & 0.943 $\pm$ 0.057 & 0.678 $\pm$ 0.099 & 0.703 $\pm$ 0.154 \\
\midrule
PUAE & 0.980 $\pm$ 0.017 & 0.942 $\pm$ 0.041 & 0.716 $\pm$ 0.053 & 0.695 $\pm$ 0.083 \\
PUSVDD & \textbf{0.994 $\pm$ 0.004} & \textbf{0.972 $\pm$ 0.029} & \textbf{0.787 $\pm$ 0.069} & \textbf{0.796 $\pm$ 0.059} \\
\bottomrule
\end{tabular}
\label{tab:seen-anomalies-simple}
\end{table}

\begin{table}[h]
\caption{
  Unseen anomaly detection performance on MNIST, FashionMNIST, SVHN, and CIFAR10.
}
\centering
\begin{tabular}{lrrrr}
\toprule
 & MNIST & FashionMNIST & SVHN & CIFAR10 \\
\midrule
IF & 0.955 $\pm$ 0.031 & \textbf{0.917 $\pm$ 0.111} & 0.492 $\pm$ 0.018 & 0.635 $\pm$ 0.095 \\
AE & 0.981 $\pm$ 0.013 & 0.842 $\pm$ 0.130 & 0.560 $\pm$ 0.045 & 0.497 $\pm$ 0.111 \\
DeepSVDD & 0.950 $\pm$ 0.044 & 0.900 $\pm$ 0.143 & 0.564 $\pm$ 0.043 & 0.724 $\pm$ 0.087 \\
LOE & 0.949 $\pm$ 0.034 & 0.892 $\pm$ 0.152 & 0.602 $\pm$ 0.058 & 0.724 $\pm$ 0.099 \\
\midrule
ABC & 0.980 $\pm$ 0.014 & 0.840 $\pm$ 0.133 & 0.559 $\pm$ 0.046 & 0.497 $\pm$ 0.109 \\
DeepSAD & 0.954 $\pm$ 0.040 & 0.900 $\pm$ 0.153 & 0.631 $\pm$ 0.044 & 0.735 $\pm$ 0.078 \\
SOEL & 0.963 $\pm$ 0.033 & 0.908 $\pm$ 0.135 & \textbf{0.702 $\pm$ 0.061} & 0.778 $\pm$ 0.087 \\
PU & 0.965 $\pm$ 0.038 & 0.901 $\pm$ 0.134 & \textbf{0.684 $\pm$ 0.102} & 0.683 $\pm$ 0.133 \\
\midrule
PUAE & \textbf{0.985 $\pm$ 0.017} & 0.894 $\pm$ 0.140 & 0.662 $\pm$ 0.078 & 0.639 $\pm$ 0.090 \\
PUSVDD & \textbf{0.983 $\pm$ 0.024} & \textbf{0.924 $\pm$ 0.134} & \textbf{0.708 $\pm$ 0.103} & \textbf{0.811 $\pm$ 0.101} \\
\bottomrule
\end{tabular}
\label{tab:unseen-anomalies-simple}
\end{table}

\begin{table}[h]
\caption{
  Seen anomaly detection performance on CIFAR100, Path, OCT and Tissue.
}
\centering
\begin{tabular}{lrrrr}
\toprule
& CIFAR100 & Path & OCT & Tissue \\
\midrule
IF & 0.577 $\pm$ 0.005 & 0.657 $\pm$ 0.211 & 0.679 $\pm$ 0.005 & 0.443 $\pm$ 0.189 \\
AE & 0.514 $\pm$ 0.013 & 0.562 $\pm$ 0.253 & \textbf{0.808 $\pm$ 0.005} & 0.443 $\pm$ 0.188 \\
DeepSVDD & 0.623 $\pm$ 0.036 & 0.769 $\pm$ 0.139 & 0.702 $\pm$ 0.043 & 0.692 $\pm$ 0.047 \\
LOE & 0.624 $\pm$ 0.035 & 0.746 $\pm$ 0.167 & 0.750 $\pm$ 0.031 & 0.657 $\pm$ 0.084 \\
\midrule
ABC & 0.516 $\pm$ 0.014 & 0.564 $\pm$ 0.252 & \textbf{0.805 $\pm$ 0.002} & 0.448 $\pm$ 0.187 \\
DeepSAD & 0.628 $\pm$ 0.042 & 0.772 $\pm$ 0.165 & \textbf{0.798 $\pm$ 0.032} & 0.715 $\pm$ 0.040 \\
SOEL & \textbf{0.696 $\pm$ 0.020} & 0.806 $\pm$ 0.142 & \textbf{0.825 $\pm$ 0.017} & 0.739 $\pm$ 0.044 \\
PU & 0.630 $\pm$ 0.023 & \textbf{0.847 $\pm$ 0.097} & 0.565 $\pm$ 0.089 & 0.628 $\pm$ 0.080 \\
\midrule
PUAE & 0.576 $\pm$ 0.025 & 0.745 $\pm$ 0.171 & 0.800 $\pm$ 0.011 & 0.613 $\pm$ 0.094 \\
PUSVDD & \textbf{0.700 $\pm$ 0.021} & \textbf{0.826 $\pm$ 0.137} & \textbf{0.822 $\pm$ 0.016} & \textbf{0.764 $\pm$ 0.044} \\
\bottomrule
\end{tabular}
\label{tab:seen-anomalies-real}
\end{table}

\begin{table}[h]
\caption{
  Unseen anomaly detection performance on CIFAR100, Path, OCT and Tissue.
}
\centering
\begin{tabular}{lrrrr}
\toprule
& CIFAR100 & Path & OCT & Tissue \\
\midrule
IF & 0.632 $\pm$ 0.005 & \textbf{0.961 $\pm$ 0.055} & 0.785 $\pm$ 0.004 & 0.500 $\pm$ 0.187 \\
AE & 0.665 $\pm$ 0.009 & 0.647 $\pm$ 0.239 & \textbf{0.965 $\pm$ 0.005} & 0.493 $\pm$ 0.170 \\
DeepSVDD & 0.552 $\pm$ 0.033 & 0.749 $\pm$ 0.200 & 0.774 $\pm$ 0.072 & 0.629 $\pm$ 0.069 \\
LOE & 0.528 $\pm$ 0.047 & 0.696 $\pm$ 0.205 & 0.851 $\pm$ 0.030 & 0.613 $\pm$ 0.093 \\
\midrule
ABC & 0.665 $\pm$ 0.009 & 0.644 $\pm$ 0.241 & \textbf{0.961 $\pm$ 0.002} & 0.496 $\pm$ 0.168 \\
DeepSAD & 0.561 $\pm$ 0.024 & 0.755 $\pm$ 0.242 & 0.874 $\pm$ 0.049 & 0.651 $\pm$ 0.075 \\
SOEL & 0.570 $\pm$ 0.034 & 0.776 $\pm$ 0.190 & 0.918 $\pm$ 0.021 & 0.666 $\pm$ 0.093 \\
PU & 0.452 $\pm$ 0.039 & 0.767 $\pm$ 0.215 & 0.712 $\pm$ 0.154 & 0.638 $\pm$ 0.098 \\
\midrule
PUAE & \textbf{0.671 $\pm$ 0.011} & 0.806 $\pm$ 0.204 & 0.941 $\pm$ 0.011 & 0.574 $\pm$ 0.116 \\
PUSVDD & 0.573 $\pm$ 0.027 & 0.835 $\pm$ 0.215 & 0.925 $\pm$ 0.022 & \textbf{0.697 $\pm$ 0.114} \\
\bottomrule
\end{tabular}
\label{tab:unseen-anomalies-real}
\end{table}

Tables~\ref{tab:seen-anomalies-simple}, \ref{tab:unseen-anomalies-simple}, \ref{tab:seen-anomalies-real} and \ref{tab:unseen-anomalies-real} show the detection performance for seen and unseen anomalies, respectively.
We show the average of the AUROC scores for all normal classes.
We used bold to highlight the best results and statistically non-different results according to a pair-wise $t$-test.
We used 5\% as the p-value.

For seen anomalies,
the PUSVDD achieved the best performance among all approaches.
These results show the effectiveness of our approach,
which is robust to the contaminated unlabeled data according to PU learning.
For unseen anomalies,
although the detection performance is highly dataset-dependent,
the PUSVDD generally performs well.
These results indicate that we may be able to improve the detection performance for unseen anomalies by using seen anomalies.

These results also show the difference between the conventional PU learning and our approach.
The PU achieved the poor detection performance for unseen anomalies.
This is because it sets the decision boundary between normal data points and seen anomalies.
On the other hand, our approach can detect unseen anomalies since it is based on the anomaly detector.

\section{Anomaly Detection Performance on MVTec AD}
\label{sec:additional_mvtec}

\begin{table}[h]
\caption{
Comparison of anomaly detection performance on MVTec AD.
}
\centering
\begin{tabular}{lrrr}
\toprule
& PUSVDD & SOEL & WSAD-DT \\
\midrule
all    & \textbf{0.793 $\pm$ 0.016} & 0.669 $\pm$ 0.021 & 0.641 $\pm$ 0.029 \\
seen   & \textbf{0.799 $\pm$ 0.013} & 0.688 $\pm$ 0.021 & 0.642 $\pm$ 0.030 \\
unseen & \textbf{0.787 $\pm$ 0.020} & 0.649 $\pm$ 0.027 & 0.640 $\pm$ 0.028 \\
\bottomrule
\end{tabular}
\label{tab:mvtec}
\end{table}

\begin{figure}[h]
  \centering
  \includegraphics[width=0.4\linewidth]{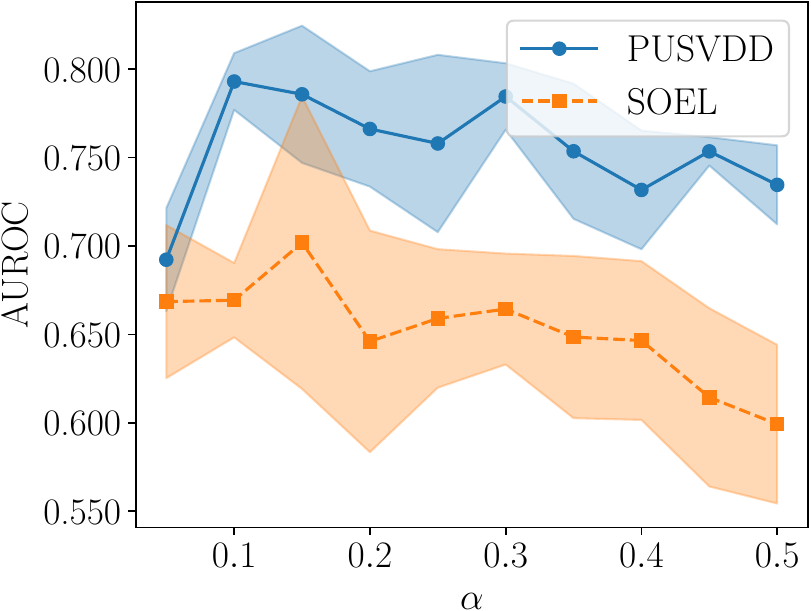}
  \caption{
    Relationship between the anomaly detection performance and the hyperparameter $\alpha$ of the PUSVDD and the SOEL on MVTec AD.
    The semi-transparent area represents standard deviations.
  }
  \label{fig:pusvdd_alpha_mvtec}
\end{figure}

As the additional experiments,
we evaluate the anomaly detection performance of the proposed method on the MVTec Anomaly Detection dataset (MVTec AD) \citep{bergmann2021mvtec,bergmann2019mvtec}.
MVTec AD is an industrial inspection dataset consisting of 5,354 high-resolution images.
The dataset contains 15 object categories, each associated with multiple types of anomalies.
In our setup, we designated the following anomaly types as unseen anomalies:
broken\_large for bottle,
bent\_wire for cable,
crack for capsule,
color for carpet,
bent for grid,
crack for hazelnut,
color for leather,
bent for metal\_nut,
color for pill,
manipulated\_front for screw,
crack for tile,
bent\_lead for transistor,
color for wood,
and broken\_teeth for zipper.
We did not set unseen anomalies for the toothbrush object because it only has one anomaly category and thus no unseen anomaly could be defined.

The remaining anomaly types were treated as seen anomalies.
Among them, 725 images were used for training and 281 images were used for test.
Out of the 725 training anomalies,
300 were used as labeled anomalies and 425 were used as unlabeled anomalies.
These 425 unlabeled anomalies were combined with 3,629 normal training images to form the unlabeled training data.

In short, the training data consisted of 4,054 unlabeled samples (3,629 normal and 425 anomaly) and 300 labeled anomaly samples.
The test data consisted of 467 normal images, 281 seen anomalies, and 252 unseen anomalies.
All images were resized to $224 \times 224$.

For the base detector,
we used the DeepSVDD that uses the pre-trained ResNet34 \citep{he2016deep} as the feature extractor.
The final fully connected layer of ResNet34 was replaced with the linear layer outputting a 128-dimensional embedding.
All affine transformations in the batch normalization layers were disabled.
We set $\alpha=0.1$ for the PUSVDD and the SOEL.
The other setups are the same as those described in experimental section.

Table~\ref{tab:mvtec} shows the comparison of AUROC among PUSVDD, SOEL, and WSAD-DT.
We used bold to highlight the best results and statistically non-different results according to a pair-wise $t$-test.
We used 5\% as the p-value.
For WSAD-DT, we used a ResNet34 encoder with the same configuration as PUSVDD and SOEL.
Figure~\ref{fig:pusvdd_alpha_mvtec} shows the relationship between the detection performance and the hyperparameter $\alpha$ of the PUSVDD and the SOEL.

The PUSVDD outperforms both the SOEL and WSAD-DT,
and the PUSVDD maintains superior performance across different values of $\alpha$.
Since WSAD-DT assumes that the unlabeled data are almost entirely normal,
it cannot handle the highly contaminated unlabeled data in our experimental setting,
resulting in lower performance.
These results indicate that our approach is also effective for high-resolution images such as MVTec AD,
and demonstrate its advantage over recent state-of-the-art semi-supervised methods under contaminated conditions.

\section{End-to-End Anomaly Detection with Class Prior Estimation}
\label{sec:alpha_estimation}

Although $\alpha$ is treated as a known hyperparameter throughout our experiments,
it can be estimated from data by combining PUSVDD with class prior estimation (CPE) \citep{christoffel2016class},
which provides a closed-form estimate of $\alpha$ in a computationally efficient manner.
The end-to-end procedure is as follows:
\begin{enumerate}
  \item Pre-train the base DeepSVDD as described in Section~\ref{sec:setup}.
  \item Extract features from the unlabeled data $\mathcal{U}$ and the anomaly data $\mathcal{A}$ using the pre-trained DeepSVDD, and apply CPE to estimate $\alpha$.
  \item Train PUSVDD using the estimated $\hat{\alpha}$.
\end{enumerate}

We applied this procedure on four datasets using class 0 as the unseen anomaly and class 1 as the normal class.
Table~\ref{tab:cpe_estimates} summarizes the resulting estimates.
On MNIST and FashionMNIST, CPE produces estimates close to the true $\alpha = 0.1$.
On SVHN and CIFAR-10, however, the estimates deviate substantially from the true value, suggesting that CPE may be unreliable on complex datasets where the feature representations of normal data and anomalies overlap considerably.
Despite this, as shown in Figure~\ref{fig:pusvdd_alpha}, PUSVDD with the estimated $\hat{\alpha}$ outperforms SOEL using the true $\alpha = 0.1$ across all four datasets, demonstrating that the performance advantage of PUSVDD is maintained even when $\alpha$ is poorly estimated.

\begin{table}[h]
\centering
\caption{CPE estimates of $\alpha$ compared with the true value ($\alpha = 0.1$).}
\label{tab:cpe_estimates}
\begin{tabular}{lc}
\toprule
Dataset & $\hat{\alpha}$ \\
\midrule
MNIST         & 0.048 \\
FashionMNIST  & 0.056 \\
SVHN          & 0.350 \\
CIFAR-10      & 0.325 \\
\bottomrule
\end{tabular}
\end{table}

\section{Anomaly Detection Performance for Each Normal Class}
\label{sec:per_class}

We provide per-class AUROC scores for PUSVDD and SOEL on MNIST, FashionMNIST, SVHN, CIFAR10, Path, and Tissue in Tables~\ref{tab:per_class_mnist}, \ref{tab:per_class_fashionmnist}, \ref{tab:per_class_svhn}, \ref{tab:per_class_cifar10}, \ref{tab:per_class_path}, and~\ref{tab:per_class_tissue}.
For each dataset, class 0 is fixed as the unseen anomaly, and each remaining class is selected as the normal class in turn; all other classes serve as seen anomalies.
We ran each experiment five times with different random seeds.
Bold indicates the best results and statistically non-different results according to a pairwise $t$-test with a significance level of 5\%.
The averages of these per-class results correspond to those reported in Tables~\ref{tab:results_first} and~\ref{tab:results_last}.

While SOEL suffers substantial performance degradation for several normal classes,
PUSVDD consistently achieves detection performance that is comparable to or better than SOEL across all normal classes,
demonstrating the strong practical utility of the proposed method.

\begin{table}[h]
\centering
\caption{Per-class AUROC on MNIST. Class 0 (digit 0) is the unseen anomaly.}
\begin{tabular}{crr}
\toprule
Class & SOEL & PUSVDD \\
\midrule
1 & 0.995 $\pm$ 0.001 & \textbf{0.999 $\pm$ 0.000} \\
2 & \textbf{0.927 $\pm$ 0.047} & \textbf{0.988 $\pm$ 0.003} \\
3 & \textbf{0.986 $\pm$ 0.009} & \textbf{0.996 $\pm$ 0.001} \\
4 & \textbf{0.982 $\pm$ 0.013} & \textbf{0.997 $\pm$ 0.001} \\
5 & 0.935 $\pm$ 0.024 & \textbf{0.989 $\pm$ 0.009} \\
6 & \textbf{0.964 $\pm$ 0.005} & \textbf{0.967 $\pm$ 0.013} \\
7 & 0.974 $\pm$ 0.006 & \textbf{0.996 $\pm$ 0.002} \\
8 & 0.939 $\pm$ 0.023 & \textbf{0.987 $\pm$ 0.004} \\
9 & 0.960 $\pm$ 0.010 & \textbf{0.987 $\pm$ 0.004} \\
\bottomrule
\end{tabular}
\label{tab:per_class_mnist}
\end{table}

\begin{table}[h]
\centering
\caption{Per-class AUROC on FashionMNIST. Class 0 (T-shirt/top) is the unseen anomaly.}
\begin{tabular}{crr}
\toprule
Class & SOEL & PUSVDD \\
\midrule
1 & 0.983 $\pm$ 0.003 & \textbf{0.993 $\pm$ 0.002} \\
2 & \textbf{0.930 $\pm$ 0.019} & \textbf{0.952 $\pm$ 0.010} \\
3 & \textbf{0.911 $\pm$ 0.007} & \textbf{0.930 $\pm$ 0.015} \\
4 & 0.920 $\pm$ 0.008 & \textbf{0.945 $\pm$ 0.014} \\
5 & 0.985 $\pm$ 0.004 & \textbf{0.993 $\pm$ 0.002} \\
6 & \textbf{0.750 $\pm$ 0.026} & \textbf{0.763 $\pm$ 0.019} \\
7 & \textbf{0.994 $\pm$ 0.001} & \textbf{0.996 $\pm$ 0.001} \\
8 & 0.960 $\pm$ 0.002 & \textbf{0.992 $\pm$ 0.001} \\
9 & 0.989 $\pm$ 0.002 & \textbf{0.996 $\pm$ 0.001} \\
\bottomrule
\end{tabular}
\label{tab:per_class_fashionmnist}
\end{table}

\begin{table}[p]
\centering
\caption{Per-class AUROC on SVHN. Class 0 (digit 0) is the unseen anomaly.}
\begin{tabular}{crr}
\toprule
Class & SOEL & PUSVDD \\
\midrule
1 & \textbf{0.783 $\pm$ 0.040} & \textbf{0.850 $\pm$ 0.048} \\
2 & 0.760 $\pm$ 0.014 & \textbf{0.801 $\pm$ 0.027} \\
3 & \textbf{0.698 $\pm$ 0.032} & \textbf{0.682 $\pm$ 0.029} \\
4 & \textbf{0.769 $\pm$ 0.033} & \textbf{0.788 $\pm$ 0.025} \\
5 & \textbf{0.765 $\pm$ 0.012} & \textbf{0.799 $\pm$ 0.022} \\
6 & \textbf{0.705 $\pm$ 0.013} & \textbf{0.690 $\pm$ 0.018} \\
7 & \textbf{0.755 $\pm$ 0.042} & \textbf{0.771 $\pm$ 0.026} \\
8 & \textbf{0.671 $\pm$ 0.022} & \textbf{0.658 $\pm$ 0.034} \\
9 & \textbf{0.694 $\pm$ 0.031} & \textbf{0.675 $\pm$ 0.014} \\
\bottomrule
\end{tabular}
\label{tab:per_class_svhn}
\end{table}

\begin{table}[p]
\centering
\caption{Per-class AUROC on CIFAR10. Class 0 (airplane) is the unseen anomaly.}
\begin{tabular}{crr}
\toprule
Class & SOEL & PUSVDD \\
\midrule
1 & \textbf{0.793 $\pm$ 0.066} & \textbf{0.824 $\pm$ 0.030} \\
2 & 0.712 $\pm$ 0.012 & \textbf{0.724 $\pm$ 0.010} \\
3 & \textbf{0.724 $\pm$ 0.054} & \textbf{0.795 $\pm$ 0.026} \\
4 & \textbf{0.731 $\pm$ 0.096} & \textbf{0.792 $\pm$ 0.015} \\
5 & 0.790 $\pm$ 0.029 & \textbf{0.843 $\pm$ 0.025} \\
6 & \textbf{0.856 $\pm$ 0.016} & \textbf{0.852 $\pm$ 0.023} \\
7 & 0.780 $\pm$ 0.032 & \textbf{0.852 $\pm$ 0.016} \\
8 & \textbf{0.751 $\pm$ 0.015} & \textbf{0.766 $\pm$ 0.018} \\
9 & \textbf{0.782 $\pm$ 0.031} & \textbf{0.805 $\pm$ 0.023} \\
\bottomrule
\end{tabular}
\label{tab:per_class_cifar10}
\end{table}

\begin{table}[p]
\centering
\caption{Per-class AUROC on Path. Class 0 is the unseen anomaly.}
\begin{tabular}{crr}
\toprule
Class & SOEL & PUSVDD \\
\midrule
1 & \textbf{0.965 $\pm$ 0.051} & \textbf{0.996 $\pm$ 0.005} \\
2 & \textbf{0.526 $\pm$ 0.126} & \textbf{0.554 $\pm$ 0.123} \\
3 & \textbf{0.950 $\pm$ 0.035} & \textbf{0.957 $\pm$ 0.007} \\
4 & \textbf{0.737 $\pm$ 0.031} & \textbf{0.733 $\pm$ 0.068} \\
5 & \textbf{0.741 $\pm$ 0.035} & \textbf{0.751 $\pm$ 0.096} \\
6 & \textbf{0.812 $\pm$ 0.082} & \textbf{0.859 $\pm$ 0.086} \\
7 & 0.803 $\pm$ 0.030 & \textbf{0.874 $\pm$ 0.023} \\
8 & 0.797 $\pm$ 0.086 & \textbf{0.922 $\pm$ 0.029} \\
\bottomrule
\end{tabular}
\label{tab:per_class_path}
\end{table}

\begin{table}[p]
\centering
\caption{Per-class AUROC on Tissue. Class 0 is the unseen anomaly.}
\begin{tabular}{crr}
\toprule
Class & SOEL & PUSVDD \\
\midrule
1 & \textbf{0.623 $\pm$ 0.017} & \textbf{0.622 $\pm$ 0.010} \\
2 & \textbf{0.788 $\pm$ 0.052} & \textbf{0.843 $\pm$ 0.029} \\
3 & 0.739 $\pm$ 0.029 & \textbf{0.806 $\pm$ 0.030} \\
4 & 0.693 $\pm$ 0.016 & \textbf{0.749 $\pm$ 0.018} \\
5 & \textbf{0.686 $\pm$ 0.035} & \textbf{0.722 $\pm$ 0.040} \\
6 & \textbf{0.739 $\pm$ 0.026} & \textbf{0.716 $\pm$ 0.011} \\
7 & \textbf{0.651 $\pm$ 0.037} & \textbf{0.655 $\pm$ 0.037} \\
\bottomrule
\end{tabular}
\label{tab:per_class_tissue}
\end{table}

\FloatBarrier
\section{Analysis of Unseen Anomaly Contamination}
\label{sec:unseen_contamination}

The approximation in Eq.~(\ref{eq:pn_loss_approx}) assumes that the unlabeled anomalies follow the same distribution $p_{\mathcal{A}}$ as the labeled anomaly data.
When the unlabeled data also contain unseen anomalies from a different distribution,
this assumption does not hold, and the unbiased risk estimator becomes biased.
As discussed in Section~\ref{sec:conclusion},
we examine this case by using the toy dataset in Figure~\ref{fig:toy},
adding unseen anomalies as a Gaussian mixture on the diagonal opposite to the seen anomalies.
We fix the contamination ratio $\alpha=0.1$ and vary $\beta$,
which is the fraction of the unlabeled anomalies drawn from the unseen distribution,
so that $\beta=0$ recovers the original setting and $\beta=1$ maximally violates the assumption.
The labeled anomaly data are drawn from the seen distribution only.
We compare PUAE with SOEL using the same DAE as the base anomaly detector,
so that only the training objective differs.

\begin{figure}[h]
  \centering
    \begin{minipage}{0.32\linewidth}
      \includegraphics[width=\linewidth]{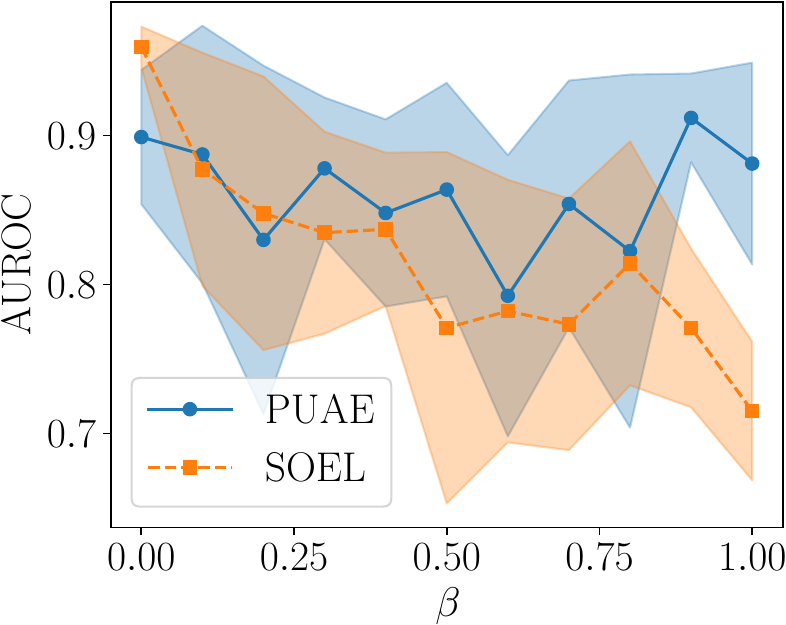}
      \subcaption{Overall}
      \label{fig:toy_unseen_overall}
    \end{minipage}
    \begin{minipage}{0.32\linewidth}
      \includegraphics[width=\linewidth]{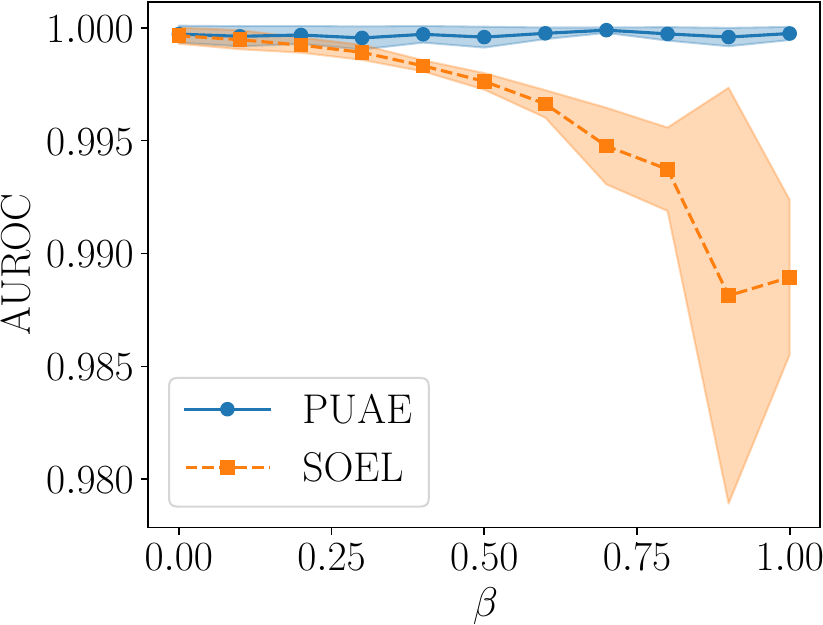}
      \subcaption{Seen}
      \label{fig:toy_unseen_seen}
    \end{minipage}
    \begin{minipage}{0.32\linewidth}
      \includegraphics[width=\linewidth]{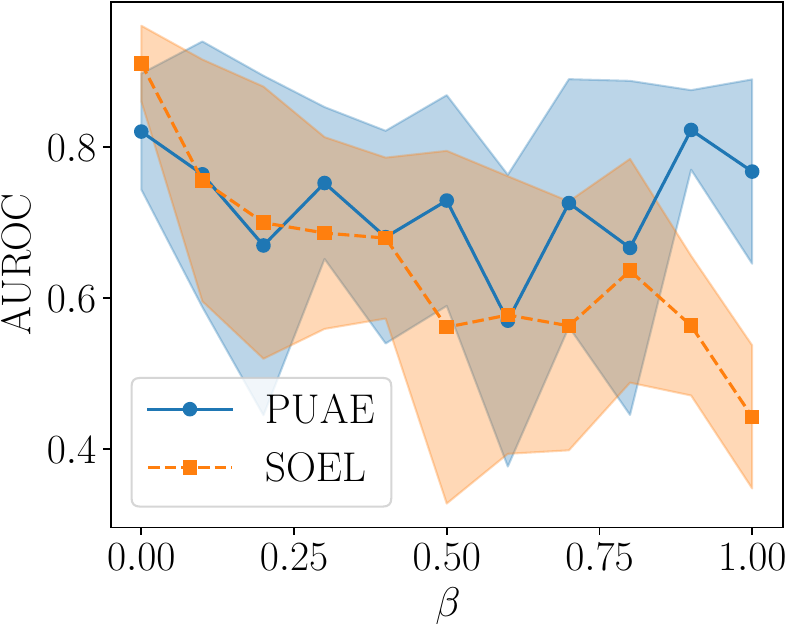}
      \subcaption{Unseen}
      \label{fig:toy_unseen_unseen}
    \end{minipage}
  \caption{
    The anomaly detection performance of the PUAE and the SOEL as the fraction $\beta$ of unseen anomalies among the unlabeled anomalies increases, with the total contamination ratio fixed at $\alpha=0.1$.
    Both methods use the same DAE as the base anomaly detector, so that only the training objective differs.
    The three panels show the AUROC on the whole test set, on the seen anomalies, and on the unseen anomalies, respectively.
    The semi-transparent area represents standard deviations.
  }
  \label{fig:toy_unseen}
\end{figure}

Figure~\ref{fig:toy_unseen} shows the AUROC of PUAE and SOEL for increasing $\beta$.
The two methods are comparable at $\beta=0$ within the standard deviations,
but as $\beta$ increases, the AUROC of SOEL clearly decreases for all cases,
while that of PUAE degrades much less,
so PUAE becomes more robust than SOEL under unseen anomaly contamination.
In particular, the AUROC of PUAE for the seen anomalies remains almost unchanged for all $\beta$.
This is because PUAE uses the labeled anomaly data through the unbiased PU risk,
and is thus unaffected by the unseen anomalies in the unlabeled data,
whereas SOEL relies on the pseudo-labels of the unlabeled data, which are disturbed by them.
Consistent with the discussion in Section~\ref{sec:conclusion},
the overall performance of PUAE remains high even at $\beta=1$,
which indicates that the proposed method is tolerant to the violation of the assumption.

\end{document}